%% file: example_paper.tex
\documentclass[twocolumn]{article} 
\usepackage{times}
\usepackage{geometry}
\usepackage[utf8]{inputenc}
\usepackage{microtype}
\usepackage{graphicx}
\usepackage{subcaption}
\usepackage{booktabs}
\usepackage{hyperref}
\usepackage{amsmath, amssymb, mathtools, amsthm}
\usepackage[capitalize,noabbrev]{cleveref}
\usepackage[textsize=tiny]{todonotes}
\usepackage{epigraph}
\usepackage{makecell}
\usepackage{multirow}
\usepackage{xcolor}
\usepackage{colortbl}
\definecolor{rowgray}{gray}{0.92} 
\usepackage{amsfonts}
\usepackage{adjustbox}
\usepackage{algorithm}
\usepackage{algorithmic}
\usepackage{tabularx}
\usepackage{fontawesome5}
\usepackage{tikz}
\usepackage{natbib}
\theoremstyle{plain}

\theoremstyle{definition}

\theoremstyle{remark}

\title{
    \raisebox{-6pt}{\includegraphics[height=1.5em]{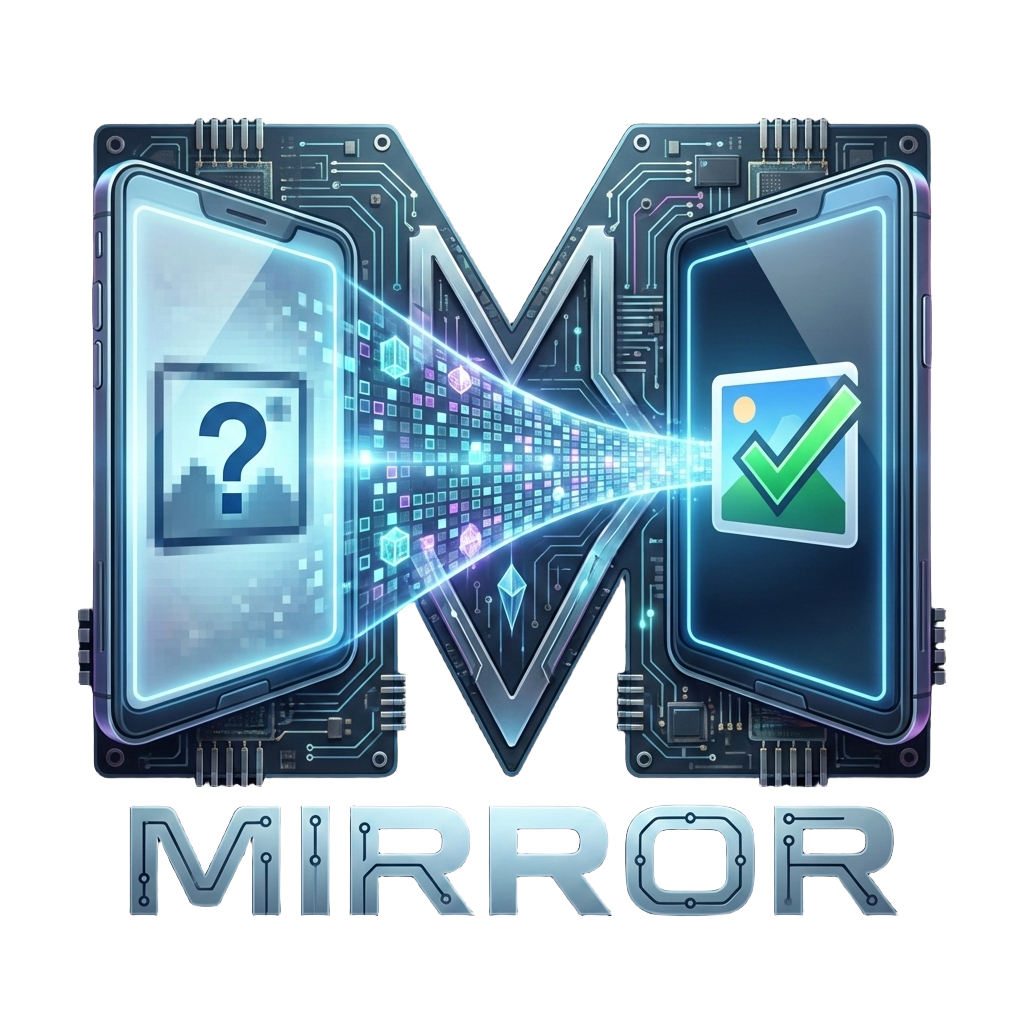}} 
    \textcolor{violet}{MIRROR}: 
    \textbf{M}anifold \textbf{I}deal \textbf{R}eference \textbf{R}econstruct\textbf{OR} for Generalizable AI-Generated Image Detection
}

\author{
    {Ruiqi Liu}$^{1,2,*}$\quad 
    {Manni Cui}$^{3,*}$ \quad
    {Ziheng Qin}$^{1,*}$ \quad
    {Zhiyuan Yan}$^{6}$ \quad
    {Ruoxin Chen}$^{4}$ \quad
    {Yi Han}$^{5}$ \quad 
    {Zhiheng Li}$^{1}$ \quad \\
    {Junkai Chen}$^{1}$ \quad 
    {ZhiJin Chen}$^{1}$ \quad
    {Kaiqing Lin}$^{8}$ \quad
    {Jialiang Shen}$^{7}$ \quad
    {Lubin Weng}$^{1}$ \quad 
    {Jing Dong}$^{1}$ \quad\\
    {Yan Wang}$^{9,\dagger}$ \quad
    {Shu Wu}$^{1,\dagger}$
    \vspace{0.3em} \\ 
    \small
    $^{1}$Institute of Automation, Chinese Academy of Sciences \quad
    $^{2}$School of Advanced Interdisciplinary Sciences, UCAS \\
    \small$^{3}$Huazhong University of Science and Technology \quad
    $^{4}$Tencent YouTu Lab \quad
    $^{5}$Southwest University \\
    \small$^{6}$Peking University \quad
    $^{7}$The University of Sydney \quad
    $^{8}$Shenzhen University \quad
    $^{9}$Tsinghua University \\ 
    \small $^*$Equal contribution \quad $^\dagger$Corresponding author
}
\date{} 

\begin{document}

\twocolumn[
    \begin{@twocolumnfalse}
        \maketitle
        \begin{center}
            \centering
            \vspace{-5mm}
            \setkeys{Gin}{width=0.95\textwidth} 
            \includegraphics{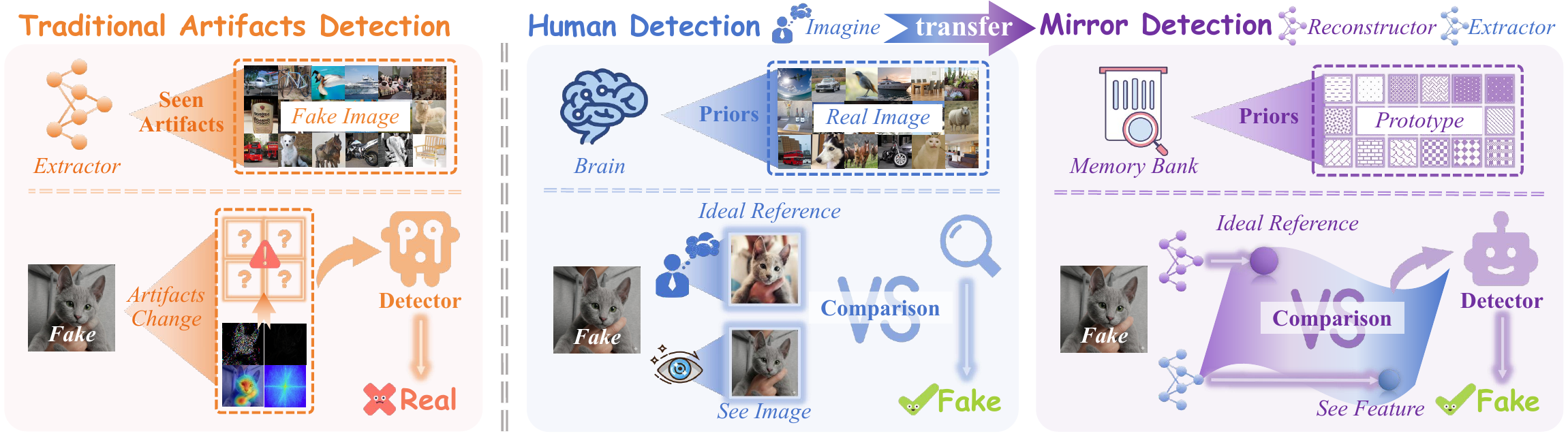}
            \captionof{figure}{\textbf{Comparison of decision paradigms for AI-generated image detection.} (Left) Traditional detectors treat detection as a Artifacts-Driven classification task, often overfitting to generator-specific artifacts. (Middle) Humans employ a \textbf{Reference-Comparison} strategy, anchoring judgment on the invariant priors of the physical world. (Right) Our \textbf{MIRROR} framework explicitly encodes these priors to act as a Reconstructor: it projects the input into a manifold-consistent \textbf{Ideal Reference}. The resulting reconstruction residual exposes forgeries as significant deviations from reality.}
            \label{fig:intro}
        \end{center}
    \end{@twocolumnfalse}
]

\input{sections/01_intro}
\input{sections/02_related_work}

\input{sections/03_method}

\input{sections/04_experiments}
\input{sections/05_conclusion}

\bibliography{references}
\bibliographystyle{plain} 

\newpage
\appendix
\onecolumn

\section{Detailed Benchmark Results and Original Weight Performance}
\label{appendix:results}

\subsection{Detailed Breakdown of Benchmark Performance}
\label{appendix:results_breakdown}

In this section, we provide the comprehensive quantitative results across five benchmarks to supplement the main experiments. Tables~\ref{tab:aigc} and \ref{tab:drct} detail the performance on \textit{AIGCDetect} and \textit{DRCT-2M}, covering a wide range of architectures from early GANs (e.g., ProGAN, StyleGAN) to modern diffusion models (e.g., SDXL). Table~\ref{tab:univfd} presents the results on \textit{UniversalFakeDetect}, focusing on robustness against unseen processing pipelines. To assess generalization across **diverse generative paradigms, ranging from open-source foundations to proprietary commercial engines**, Table~\ref{tab:wild} and Table~\ref{tab:chain} illustrate the detection accuracy on \textit{Synthbuster} (including Midjourney v6 and DALL-E 3) and the newly curated \textit{EvalGEN} benchmark, respectively. These detailed breakdowns confirm that MIRROR maintains consistent performance across diverse generative semantics and architectural paradigms, identifying forgery traces even in challenging high-fidelity samples where baseline methods often exhibit significant volatility.

\subsection{Evaluation of Official Weights}
\label{appendix:weights}

To ensure a fair comparison of architectural performance, we evaluate baseline methods using both the officially released weights (``Official'') and weights retrained on our format-aligned SDv1.4 dataset (``Ours''). Results in Table~\ref{tab:weight_comparison} present the performance under both settings. By utilizing the retrained weights, we prioritize \textbf{format alignment and data scale consistency}, ensuring that all detectors are trained under the premise of eliminating format discrepancies and maintaining identical dataset scale. This standardization eliminates discrepancies arising from varying training data sources, thereby establishing a fair and balanced baseline for analyzing intrinsic architectural capabilities.

\section{Robustness to Common Corruptions}
\label{appendix:robustness}

\begin{figure*}[h]
    \centering
    \includegraphics[width=\textwidth]{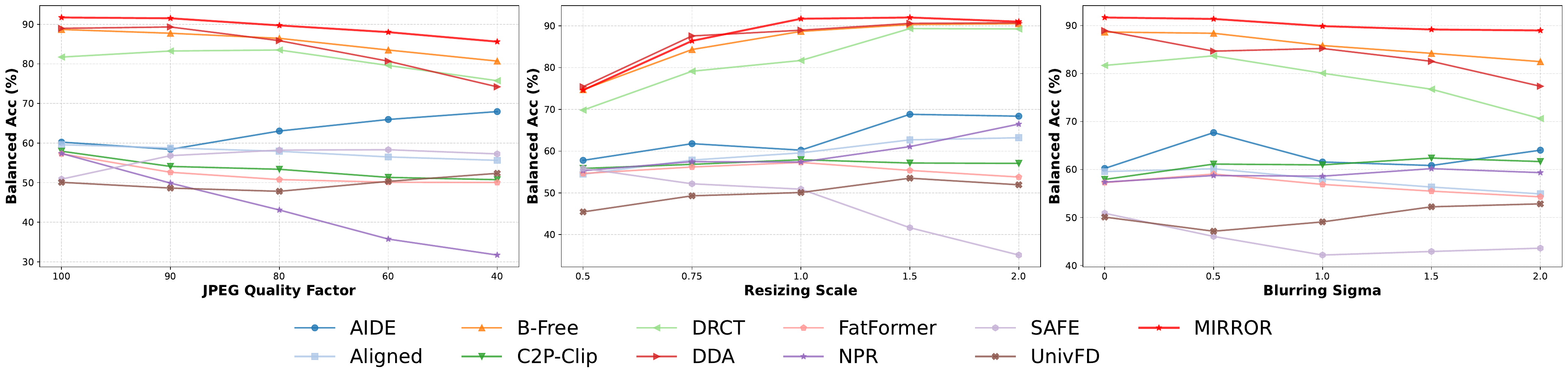}
    \caption{\textbf{Robustness evaluation under common image degradations.} We report the Balanced Accuracy curves under (Left) JPEG compression with quality factors ranging from 100 down to 40, (Middle) Spatial resizing with scale factors from 0.5$\times$ to 2.0$\times$, and (Right) Gaussian blurring with sigma values ($\sigma$) from 0 to 2.0. MIRROR (red curve) maintains high stability across all settings, whereas baseline methods exhibit significant volatility.}
    \label{fig:robustness}
\end{figure*}

To evaluate the reliability of MIRROR in real-world deployment, we conduct stress tests against three prevalent image degradations: JPEG compression, spatial resizing, and Gaussian blurring. As illustrated in Figure~\ref{fig:robustness}, MIRROR exhibits remarkable resilience across all corruption types, consistently outperforming state-of-the-art baselines. Unlike traditional artifact-driven detectors—which often suffer from severe performance degradation due to their over-reliance on fragile high-frequency patterns—MIRROR leverages the \textit{ideal reference} to maintain stable detection, even when low-level spectral cues are compromised by unseen perturbations.

\paragraph{JPEG Compression.} We simulate lossy compression artifacts by varying the JPEG quality factor from 100 down to 40. This process introduces quantization noise that effectively obliterates the subtle high-frequency spectral fingerprints often exploited by standard classifiers. MIRROR maintains its efficacy in this regime by anchoring its decision on manifold-consistent structural residuals rather than volatile pixel-level statistics, which are easily corrupted by quantization.

\paragraph{Resizing Operations.} To model resolution fluctuations typical in social media transmission, we resize images with scale factors ranging from $0.5\times$ to $2.0\times$. Such geometric transformations introduce interpolation artifacts (e.g., aliasing) that often disrupt local texture analysis. Our results confirm that MIRROR’s sparse linear combination strategy effectively preserves structural integrity, rendering the detection process largely invariant to scale shifts and preserving high-level semantic consistency.

\paragraph{Gaussian Blurring.} We apply Gaussian blurring with kernels of $\sigma \in [0, 2.0]$ to simulate out-of-focus or low-quality acquisition conditions. Blurring acts as a low-pass filter, suppressing the high-frequency components that many baselines (e.g., FatFormer, SAFE) predominantly rely on. The stability of MIRROR under these conditions underscores its capability to identify forgeries based on global semantic discordance rather than relying on fragile local sharpness cues.

\section{Heatmap Visualization}
\label{appendix:heatmap}

To qualitatively elucidate how MIRROR distinguishes authentic content from synthetic artifacts, we visualize the reconstruction residuals across various generative architectures. Unlike conventional interpretability tools such as Grad-CAM, which rely on backpropagated classification logits to identify salient regions, MIRROR derives heatmaps directly from the residual map $R = \|x - \hat{x}\|_2$. Here, $x$ denotes the input feature and $\hat{x}$ represents its corresponding manifold projection. This residual-based visualization provides a more granular forensic perspective, highlighting precisely where the generative distribution deviates from the learned manifold of authentic images.

\begin{figure*}[h]
    \centering

    \includegraphics[width=0.8\textwidth]{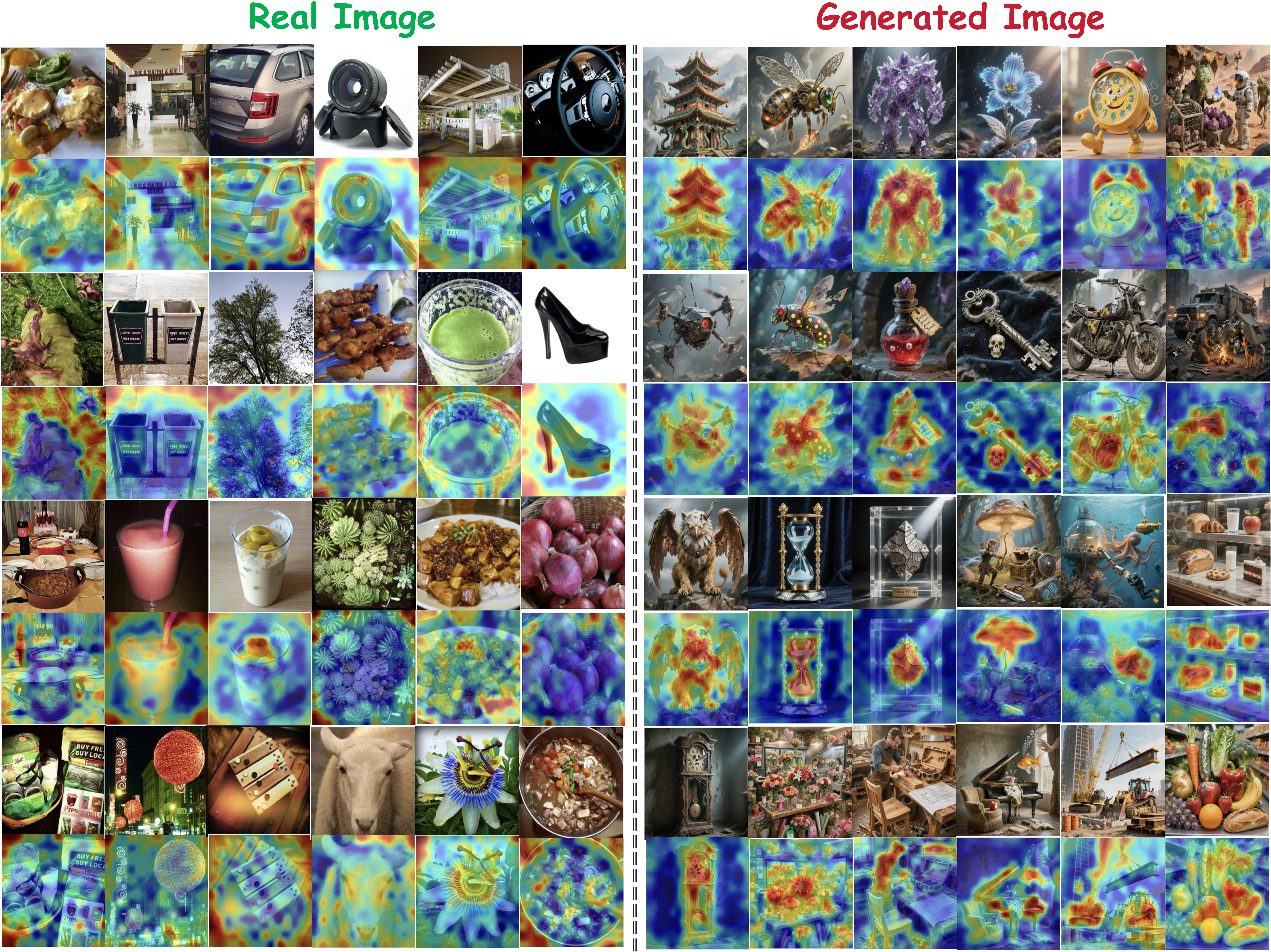} 
    \caption{\textbf{Reconstruction residual heatmaps for authentic and synthetic images.} MIRROR yields negligible residuals for authentic samples (left), signaling successful manifold projection. For AI-generated content (right), the localized high-intensity regions (red) unveil forensic inconsistencies and structural discrepancies, providing an intuitive interpretation of MIRROR’s discriminatory mechanism.}
    \label{fig:heatmap}
\end{figure*}

As illustrated in Figure~\ref{fig:heatmap}, authentic images manifest uniformly low residuals across their spatial extent, validating their close adherence to the learned manifold of natural images. Conversely, AI-generated images exhibit localized clusters of high residuals, which precisely pinpoint subtle forensic inconsistencies such as irregular illumination, blurred texture boundaries, or geometric distortions. These heatmaps demonstrate that MIRROR can effectively localize generative artifacts often imperceptible to human observers, thereby providing a transparent and interpretable foundation for its classification decisions.

\section{Visual Showcase of the Human-AIGI Benchmark} \label{appendix:dataset}
This appendix provides a visual comprehensive overview of the Human-AIGI Benchmark,  a dataset specifically curated to reflect the increasingly sophisticated threat landscape posed by modern commercial generative AI systems. Unlike traditional benchmarks that often focus on single-generator settings or rely on easily distinguishable artifacts, our collection emphasizes cross-generator heterogeneity and high perceptual indistinguishability across samples. The dataset incorporates outputs from multiple state-of-the-art generative models and diverse content domains, ensuring realistic and challenging evaluation scenarios. As such, the Human-AIGI Benchmark serves as a rigorous and representative testbed for assessing the robustness, generalization, and real-world applicability of next-generation human–AI content detectors.

\begin{figure*}[h]
\centering
\includegraphics[width=0.8\textwidth]{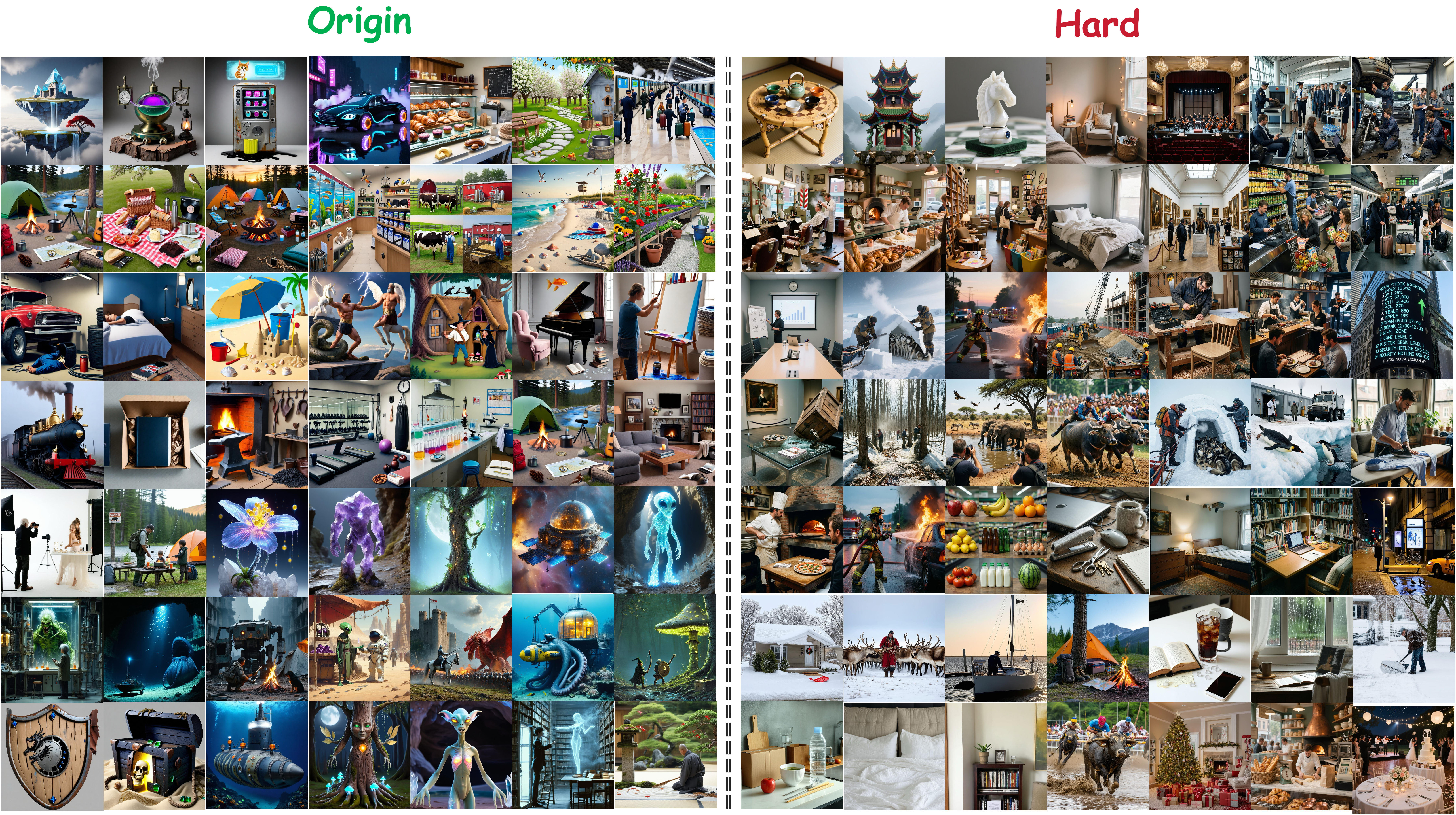}
\caption{\textbf{Visual showcase of the Human-AIGI Benchmark.} 
(Left) \textbf{Representative Samples:} A diverse collection encompassing portraits, objects, and complex scenes curated from 27 state-of-the-art generators in T2I-COREBENCH (e.g., Flux, Hunyuan 3.0, Nanobanana). These samples reflect the broad semantic coverage and high fidelity of current commercial-grade AIGI. 
(Right) \textbf{The Human-Imperceptible Subset ($\mathcal{D}_{hard}$):} A rigorous subset consisting of "perceptual corner cases" identified through psychophysical trials with 50 participants. Following the criteria $S(x) \ge \tau_{real} \lor RT(x) > \mu_{rt} + \sigma_{rt}$, these images either successfully deceived human observers into a "Real" judgment (high confidence score $S$) or induced significant cognitive hesitation (prolonged response time $RT$). This subset serves as a stringent testbed for evaluating whether a detector can identify generative threats that evade biological perceptual scrutiny and human expert review.}
\label{fig:dataset_showcase}
\end{figure*}


\begin{table*}[t!]
\centering
\vspace{1mm}
\caption{Performance comparison on the \textit{AIGCDetect} benchmark. The test benchmark uses balance accuracy (B.Acc). Bold numbers indicate the best performance in each column, and underlined numbers indicate the second-best performance.}\vspace{-2mm}
\fontsize{9pt}{11pt}\selectfont   
\setlength{\tabcolsep}{5pt}
\renewcommand{\arraystretch}{1}
\begin{adjustbox}{max width=\textwidth}
\begin{tabular}{lccccccccccccccccccccc}
\toprule
\textbf{Method} & 
\makecell{ADM} & 
\makecell{DALLE2} & 
\makecell{GLIDE} & 
\makecell{Midj.} & 
\makecell{VQDM} & 
\makecell{Big-\\GAN} & 
\makecell{Cycle-\\GAN} & 
\makecell{Gau-\\GAN} & 
\makecell{Pro-\\GAN} & 
\makecell{SDXL} & 
\makecell{SD1.4} & 
\makecell{SD1.5} & 
\makecell{Star-\\GAN} & 
\makecell{Style-\\GAN} & 
\makecell{Style-\\GAN2} & 
\makecell{WFR} & 
\makecell{Wukong} & 
\textbf{\makecell{Avg\\B.Acc}}  \\
\midrule
NPR \cite{tan2024rethinking}   & 91.4 & 43.8 & 64.0 & 83.8 & 84.1 & 50.6 & 47.7 & 55.4 & 52.1 & 63.1 & 77.5 & 77.5 & 52.8 & 53.4 & 61.2 & 51.8 & 75.8 & 63.9 \\
UnivFD \cite{ojha2023towards}  & 68.5 & 75.0 & 69.2 & 71.0 & 55.5 & 39.5 & 43.5 & 36.6 & 42.1 & 73.9 & 69.2 & 69.4 & 47.1 & 43.5 & 39.2 & 51.2 & 65.6 & 56.5 \\
FatFormer \cite{liu2024forgery}& 94.7 & 52.4 & 75.7 & 52.4 & 77.5 & 99.0 & 99.5 & 99.7 & 99.7 & 56.7 & 65.4 & 65.8 & 99.9 & 94.6 & 91.9 & 99.5 & 71.8 & 82.1 \\
SAFE \cite{li2024improving}    & 46.8 & 46.6 & 46.7 & 47.2 & 48.7 & 50.1 & 48.3 & 49.7 & 50.0 & 46.9 & 47.0 & 47.0 & 50.0 & 50.0 & 49.6 & 50.0 & 47.6 & 48.4 \\
C2P-CLIP \cite{tan2025c2p}     & 87.9 & 50.8 & 71.2 & 55.1 & 74.7 & 97.4 & 99.5 & 99.7 & 99.8 & 56.6 & 68.8 & 69.4 & 99.8 & 91.5 & 80.6 & 84.1 & 73.3 & 80.0 \\
AIDE \cite{yan2024sanity}      & 98.9 & 61.1 & 75.6 & 82.7 & 96.3 & 91.1 & 89.1 & 95.9 & 80.2 & 78.4 & 93.0 & 93.2 & 75.2 & 76.6 & 76.6 & 74.3 & 89.7 & 84.0 \\
DRCT \cite{chen2024drct}       & 53.6 & 83.0 & 61.6 & 97.7 & 64.5 & 48.8 & 48.8 & 49.2 & 50.2 & 96.8 & 98.6 & 98.5 & 43.2 & 48.5 & 49.4 & 50.0 & 98.0 & 67.1 \\
Aligned \cite{rajan2024aligned}& 60.4 & 60.9 & 65.7 & 58.2 & 60.5 & 53.0 & 50.5 & 53.1 & 51.2 & 63.2 & 54.3 & 54.8 & 49.1 & 56.7 & 50.1 & 48.9 & 56.5 & 55.7 \\
B-Free \cite{guillaro2025bias} & 76.4 & 74.9 & 70.8 & 94.3 & 89.2 & 91.5 & 66.7 & 96.1 & 95.4 & 99.6 & 99.4 & 99.3 & 81.0 & 75.4 & 71.0 & 58.9 & 99.4 & 84.7 \\
DDA \cite{chen2025dual}        & 90.7 & 93.9 & 88.8 & 93.9 & 67.8 & 81.1 & 64.8 & 81.6 & 73.1 & 98.0 & 96.7 & 96.7 & 67.9 & 67.8 & 76.7 & 49.8 & 96.8 & 81.5 \\
\midrule
\rowcolor{blue!5}
\textbf{MIRROR}     & 75.6 & 88.9 & 93.1 & 92.5 & 95.9 & 97.2 & 73.3 & 95.8 & 96.8 & 99.8 & 99.9 & 99.8 & 87.6 & 91.9 &  91.1 &78.9 & 99.8 & \textbf{91.7} \\

\bottomrule
\end{tabular}
\end{adjustbox}
\vspace{-1mm}
\label{tab:aigc}
\end{table*}

\begin{table*}[t!]
\centering
\caption{Performance comparison on the \textit{DRCT-2M} benchmark.}
\vspace{-2mm}
\fontsize{9pt}{11pt}\selectfont
\setlength{\tabcolsep}{5pt}
\renewcommand{\arraystretch}{1}
\begin{adjustbox}{max width=\textwidth}
\begin{tabular}{lccccccccccccccccc}
\toprule
\textbf{Method} &
LDM &
SDv1.4 &
SDv1.5 &
SDv2 &
SDXL &
\makecell{SDXL-\\Refiner} &
\makecell{SD-\\Turbo} &
\makecell{SDXL-\\Turbo} &
\makecell{LCM-\\SDv1.5} &
\makecell{LCM-\\SDXL} &
\makecell{SDv1-\\Ctrl} &
\makecell{SDv2-\\Ctrl} &
\makecell{SDXL-\\Ctrl} &
\makecell{SDv1-\\DR} &
\makecell{SDv2-\\DR} &
\makecell{SDXL-\\DR} &
\textbf{\makecell{Avg\\B.Acc}} \\
\midrule
NPR \cite{tan2024rethinking} & 48.8 & 59.9 & 59.6 & 62.3 & 68.7 & 74.6 & 72.7 & 63.7 & 62.5 & 73.3 & 55.8 & 53.0 & 46.0 & 50.9 & 50.1 & 50.4 & 59.5 \\
UnivFD \cite{ojha2023towards} & 66.9 & 73.0 & 73.0 & 73.9 & 72.4 & 73.4 & 73.7 & 74.3 & 74.4 & 73.3 & 68.5 & 72.4 & 70.2 & 53.3 & 58.3 & 61.2 & 69.5 \\
FatFormer \cite{liu2024forgery} & 54.6 & 50.3 & 50.1 & 50.0 & 50.5 & 49.9 & 50.0 & 49.9 & 50.6 & 53.2 & 52.3 & 53.0 & 65.4 & 70.8 & 59.6 & 51.9 & 53.9 \\
SAFE \cite{li2024improving} & 50.0 & 50.0 & 50.0 & 49.9 & 49.9 & 49.9 & 49.9 & 49.9 & 50.1 & 49.9 & 51.1 & 50.1 & 50.2 & 52.1 & 50.3 & 51.6 & 50.3 \\
C2P-CLIP \cite{tan2025c2p} & 59.6 & 51.1 & 50.9 & 50.5 & 52.2 & 50.8 & 50.6 & 50.3 & 50.4 & 53.1 & 55.6 & 53.2 & 68.6 & 66.6 & 54.9 & 52.8 & 54.4 \\
AIDE \cite{yan2024sanity} & 58.7 & 60.0 & 60.0 & 56.8 & 51.9 & 66.0 & 51.3 & 50.7 & 61.8 & 56.6 & 55.3 & 52.2 & 51.5 & 83.1 & 72.1 & 60.1 & 59.2 \\
DRCT \cite{chen2024drct} & 99.8 & 97.3 & 97.8 & 99.3 & 96.8 & 96.8 & 99.3 & 87.3 & 99.3 & 95.3 & 95.3 & 99.8 & 96.8 & 96.3 & 99.8 & 94.3 & 96.9 \\
Aligned \cite{rajan2024aligned} & 51.1 & 53.7 & 53.8 & 51.9 & 57.9 & 63.1 & 51.7 & 57.2 & 58.1 & 69.1 & 50.8 & 52.5 & 50.8 & 52.9 & 51.9 & 51.3 & 54.9 \\
B-Free \cite{guillaro2025bias}  & 99.7 & 99.7 & 99.7 & 99.6 & 99.1 & 99.4 & 99.2 & 98.3 & 99.6 & 99.6 & 99.8 & 99.8 & 99.8 & 99.5 & 99.6 & 95.4 & \textbf{99.2} \\
DDA \cite{chen2025dual} & 97.6 & 97.4 & 97.4 & 97.3 & 97.2 & 94.4 & 97.4 & 94.1 & 95.5 & 97.4 & 97.7 & 97.9 & 97.9 & 97.9 & 98.1 & 97.5 & 97.0 \\
\midrule
\rowcolor{blue!5}
\textbf{MIRROR} & 99.8 & 99.6 & 99.6 & 99.5 & 91.7 & 99.5 & 99.4 & 96.9 & 99.9 & 97.0 & 99.8 & 99.0 & 98.7 & 80.9 & 67.9 & 59.4 & 93.0 \\
\bottomrule
\end{tabular}
\end{adjustbox}
\vspace{-1mm}
\label{tab:drct}
\end{table*}

\begin{table*}[t!]
\centering
\caption{Performance comparison on the \textit{UniversalFakeDetect} benchmark.}
\vspace{-2mm}
\fontsize{9pt}{11pt}\selectfont
\setlength{\tabcolsep}{3.5pt}
\renewcommand{\arraystretch}{1}
\begin{adjustbox}{max width=\textwidth}
\begin{tabular}{lccccccccccccc}
\toprule
\textbf{Method} &
SITD &
SAN &
CRN &
IMLE &
Guided &
\makecell{LDM-\\200} &
\makecell{LDM-\\200\_cfg} &
\makecell{LDM-\\100} &
\makecell{Glide-\\100\_27}&
\makecell{Glide-\\50\_27}&
\makecell{Glide-\\100\_10}&
Dalle &
\textbf{\makecell{Avg\\B.Acc}} \\
\midrule
NPR \cite{tan2024rethinking} & 57.8 & 36.8 & 52.7 & 52.6 & 78.5 & 50.4 & 50.4 & 49.6 & 46.9 & 50.9 & 44.2 & 43.3 & 51.1 \\
UnivFD \cite{ojha2023towards} & 54.7 & 46.8 & 23.0 & 36.6 & 53.5 & 44.0 & 45.3 & 43.7 & 61.6 & 62.2 & 62.1 & 46.1 & 48.3 \\
FatFormer \cite{liu2024forgery} & 90.6 & 75.3 & 93.4 & 93.4 & 65.7 & 95.8 & 83.4 & 95.8 & 84.8 & 83.6 & 83.2 & 95.8 & 86.7 \\
SAFE \cite{li2024improving} & 50.0 & 47.0 & 50.0 & 50.0 & 48.2 & 50.7 & 50.3 & 50.3 & 49.9 & 49.9 & 49.9 & 50.4 & 49.7 \\
C2P-CLIP \cite{tan2025c2p} & 91.7 & 80.8 & 95.1 & 95.1 & 60.2 & 97.0 & 84.0 & 96.8 & 84.2 & 83.4 & 84.9 & 94.4 & 87.3 \\
AIDE \cite{yan2024sanity} & 55.0 & 60.5 & 53.1 & 76.4 & 87.4 & 93.5 & 85.6 & 93.7 & 78.3 & 78.1 & 81.1 & 91.7 & 77.9 \\
DRCT \cite{chen2024drct} & 78.3 & 82.0 & 31.1 & 48.2 & 62.5 & 90.6 & 91.4 & 90.5 & 55.6 & 54.5 & 58.8 & 57.9 & 66.8 \\
Aligned \cite{rajan2024aligned} & 63.9 & 54.1 & 54.5 & 68.9 & 57.1 & 51.0 & 55.0 & 51.4 & 68.8 & 70.5 & 68.4 & 62.3 & 60.5 \\
DDA \cite{chen2025dual} & 48.9 & 87.0 & 60.8 & 67.5 & 88.1 & 73.7 & 73.5 & 72.9 & 69.1 & 67.9 & 71.6 & 51.7 & 69.4 \\
B-Free \cite{guillaro2025bias} & 76.4 & 92.9 & 90.4 & 94.2 & 75.5 & 97.1 & 97.1 & 97.2 & 77.9 & 79.2 & 82.0 & 93.6 & 87.8 \\
\midrule
\rowcolor{blue!5}
\textbf{MIRROR} & 63.3 & 80.6 & 76.4 & 76.9 & 79.8 & 98.9 & 98.8 & 99.0 & 95.1 & 94.7 & 95.6 & 98.6 & \textbf{88.2} \\
\bottomrule
\end{tabular}
\end{adjustbox}
\vspace{-1mm}
\label{tab:univfd}
\end{table*}

\begin{table*}[!t]
\setlength{\tabcolsep}{4pt}
\renewcommand{\arraystretch}{1.2}
\centering
\begin{minipage}[t]{0.58\linewidth}
\raggedright
\caption{Performance comparison on the \textit{Synthbuster} benchmark.}\vspace{-2mm}

\begin{adjustbox}{width=\linewidth}
\begin{tabular}{lcccccccccc}
\toprule
\textbf{Method} &
DALLE2 &
DALLE3 &
Firefly &
GLIDE &
Midj. &
SD1.3 &
SD1.4 &
SD2 &
SDXL &
\textbf{\makecell{Avg\\B.Acc}} \\
\midrule
NPR \cite{tan2024rethinking} & 72.3 & 70.2 & 74.0 & 26.8 & 77.9 & 68.3 & 69.0 & 87.7 & 77.5 & 69.3 \\
UnivFD \cite{ojha2023towards} & 88.7 & 75.9 & 74.9 & 89.1 & 84.2 & 71.2 & 66.2 & 75.4 & 87.0 & 79.2 \\
FatFormer \cite{liu2024forgery} & 32.5 & 0.4 & 80.7 & 43.9 & 5.9 & 57.6 & 56.7 & 27.9 & 40.7 & 38.5 \\
SAFE \cite{li2024improving} & 0.5 & 2.5 & 0.6 & 0.0 & 6.4 & 9.5 & 7.6 & 1.8 & 0.4 & 3.3 \\
C2P-CLIP \cite{tan2025c2p} & 25.2 & 0.6 & 78.1 & 39.4 & 10.0 & 54.7 & 54.5 & 44.5 & 44.4 & 39.0 \\
AIDE \cite{yan2024sanity} & 16.5 & 23.6 & 0.3 & 42.1 & 51.5 & 88.7 & 90.2 & 77.4 & 67.3 & 50.8 \\
DRCT \cite{chen2024drct} & 3.9 & 36.8 & 13.5 & 21.6 & 98.9 & 94.9 & 93.8 & 99.9 & 96.4 & 62.2 \\
Aligned \cite{rajan2024aligned} & 23.7 & 2.3 & 28.0 & 22.0 & 13.1 & 4.2 & 3.4 & 4.6 & 14.9 & 12.9 \\
DDA \cite{chen2025dual} & 80.7 & 92.5 & 95.3 & 84.3 & 100.0 & 99.5 & 99.5 & 99.8 & 100.0 & 94.6 \\
B-Free \cite{guillaro2025bias} & 89.9 & 93.7 & 99.2 & 45.8 & 98.8 & 100.0 & 99.8 & 99.5 & 99.9 & 91.8 \\
\midrule
\rowcolor{blue!5}
\textbf{MIRROR} & 98.0 & 99.9 & 91.6 &96.5 & 98.3 & 99.8 & 100.0& 99.2 & 99.5 & \textbf{98.1} \\
\bottomrule
\end{tabular}
\end{adjustbox}
\label{tab:wild}
\end{minipage}%
\hfill
\begin{minipage}[t]{0.40\linewidth}
\raggedleft
\caption{Performance comparison on the \textit{EvalGEN}.}\vspace{-2mm}
\begin{adjustbox}{width=\linewidth}
\begin{tabular}{lcccccc}
\toprule
\textbf{Method} &
Flux &
GoT &
Infinity &
NOVA &
OmiGen &
\textbf{\makecell{Avg\\B.Acc}} \\
\midrule
NPR \cite{tan2024rethinking} & 74.4 & 70.5 & 69.7 & 76.0 & 70.7 & 72.3 \\
UnivFD \cite{ojha2023towards} & 94.3 & 96.6 & 91.9 & 95.5 & 98.1 & 95.3 \\
FatFormer \cite{liu2024forgery} & 0.4 & 2.6 & 8.6 & 54.2 & 1.7 & 13.5 \\
SAFE \cite{li2024improving} & 0.3 & 0.3 & 0.9 & 0.2 & 0.6 & 0.5 \\
C2P-CLIP \cite{tan2025c2p} & 1.8 & 6.7 & 11.4 & 46.6 & 2.7 & 13.8 \\
AIDE \cite{yan2024sanity} & 17.9 & 24.7 & 3.4 & 16.3 & 33.4 & 19.1 \\
DRCT \cite{chen2024drct} & 52.2 & 71.2 & 58.3 & 67.9 & 70.5 & 64.0 \\
Aligned \cite{rajan2024aligned} & 30.0 & 46.2 & 20.8 & 47.0 & 32.9 & 35.4 \\
DDA \cite{chen2025dual} & 89.1 & 94.9 & 95.0 & 98.3 & 96.2 & 94.7 \\
B-Free \cite{guillaro2025bias} & 52.1 & 99.5 & 97.1 & 99.9 & 99.4 & 89.6 \\
\midrule
\rowcolor{blue!5}
\textbf{MIRROR} & 99.9 & 99.8 & 99.5 & 99.9 & 95.7 & \textbf{99.0} \\
\bottomrule
\end{tabular}
\end{adjustbox}
\label{tab:chain}
\end{minipage}
\vspace{-3mm}
\end{table*}

\begin{table*}[t!]
\centering
\caption{Performance comparison between \textbf{Official Weights} and \textbf{Our Retrained Weights}. We report Balanced Accuracy (B.Acc), JPEG Robustness (J.Rob), and Resize Robustness (R.Rob). For \textbf{Synthbuster} and \textbf{EvalGEN}, which only contain fake images, the reported scores are the average of the dataset performance and the performance on the \textbf{MSCOCO} set. ``Official'' denotes results using officially released weights, and ``Ours'' denotes results obtained by training on our dataset.}
\vspace{-2mm}
\fontsize{9pt}{11pt}\selectfont
\setlength{\tabcolsep}{4pt}
\renewcommand{\arraystretch}{1.1}
\begin{adjustbox}{max width=\textwidth}
\begin{tabular}{l | c | c c c | c c c | c c c | c c c | c c c | c c c | c c c}
\toprule
\multirow{2}{*}{\textbf{Method}} & \multirow{2}{*}{\textbf{Weights}} & 
\multicolumn{3}{c|}{\textbf{AIGCDetect}} & 
\multicolumn{3}{c|}{\textbf{GenImage}} & 
\multicolumn{3}{c|}{\textbf{DRCT-2M}} & 
\multicolumn{3}{c|}{\textbf{UnivFakeDetect}} & 
\multicolumn{3}{c|}{\textbf{Synthbuster}} & 
\multicolumn{3}{c|}{\textbf{EvalGEN}} & 
\multicolumn{3}{c}{\textbf{Average}} \\
\cmidrule(lr){3-5} \cmidrule(lr){6-8} \cmidrule(lr){9-11} \cmidrule(lr){12-14} \cmidrule(lr){15-17} \cmidrule(lr){18-20} \cmidrule(lr){21-23}
 & & B.Acc & J.Rob & R.Rob & B.Acc & J.Rob & R.Rob & B.Acc & J.Rob & R.Rob & B.Acc & J.Rob & R.Rob & B.Acc & J.Rob & R.Rob & B.Acc & J.Rob & R.Rob & B.Acc & J.Rob & R.Rob \\
\midrule

\multirow{2}{*}{AIDE~\cite{yan2024sanity}} 
& Official & 81.8 & 53.7 & 75.5 & 88.4 & 57.3 & 77.3 & 66.7 & 59.8 & 63.6 & 81.1 & 51.8 & 72.8 & 80.6 & 54.0 & 78.3 & 59.7 & 59.9 & 69.8 & 76.4 & 56.1 & 72.9 \\
& Ours     & 84.0 & 53.4 & 77.0 & 88.6 & 55.3 & 76.3 & 59.2 & 54.6 & 72.5 & 77.9 & 51.5 & 70.8 &75.4 & 76.8 & 62.5 & 59.5 & 58.8 & 76.4 & 74.1 & 58.4 & 72.6 \\
\midrule

\multirow{2}{*}{FatFormer~\cite{liu2024forgery}} 
& Official & 51.4 & 51.1 & 52.0 & 52.2 & 51.3 & 52.6 & 53.0 & 51.6 & 53.5 & 52.5 & 52.1 & 52.5 & 53.9 & 52.5 & 54.6 & 54.3 & 52.8 & 54.9 & 52.9 & 51.9 & 53.4 \\
& Ours     & 82.1 & 71.5 & 80.0 & 71.5 & 58.4 & 68.4 & 53.9 & 52.1 & 52.4 & 86.7 & 68.9 & 85.2 & 69.2 & 57.1 & 65.6 & 56.7 & 53.4 & 57.5 & 70.0 & 60.2 & 68.2 \\
\midrule

\multirow{2}{*}{NPR~\cite{tan2024rethinking}} 
& Official & 72.4 & 60.0 & 68.8 & 84.7 & 67.9 & 74.9 & 70.3 & 64.7 & 63.1 & 64.2 & 56.2 & 58.0 & 82.6 & 63.3 & 76.2 & 82.9 & 80.5 & 77.9 & 76.2 & 65.4 & 69.8 \\
& Ours     & 63.9 & 64.0 & 48.2 & 73.7 & 73.7 & 46.6 & 59.5 & 58.3 & 37.4 & 51.1 & 51.1 & 49.2 & 64.6 & 64.0 & 34.4 & 66.1 & 65.7 & 64.2 & 63.2 & 62.8 & 46.7 \\
\midrule

\multirow{2}{*}{SAFE~\cite{li2024improving}} 
& Official & 49.4 & 50.1 & 35.3 & 50.6 & 50.1 & 34.4 & 49.0 & 49.9 & 44.8 & 61.4 & 49.8 & 46.5 & 48.9 & 50.0 & 25.7 & 51.4 & 50.0 & 43.9 & 51.8 & 50.0 & 38.4 \\
& Ours    & 48.4 & 57.8 & 49.4 & 47.7 & 64.4 & 49.4 & 50.3 & 62.4 & 49.2 & 49.7 & 48.5 & 49.8 & 51.6 & 53.8 & 49.9 & 50.2 & 55.4 & 49.1 & 49.7 & 57.1 & 49.5 \\
\midrule

\multirow{2}{*}{UnivFD~\cite{ojha2023towards}} 
& Official & 76.3 & 71.0 & 73.6 & 67.2 & 62.5 & 65.2 & 62.1 & 61.9 & 58.7 & 76.0 & 70.4 & 72.7 & 68.5 & 64.6 & 65.0 & 58.2 & 57.7 & 55.9 & 68.1 & 64.7 & 65.2 \\
& Ours     & 56.5 & 50.1 & 48.7 & 62.5 & 53.6 & 50.8 & 69.5 & 57.3 & 52.5 & 48.3 & 51.1 & 43.0 & 65.8 & 59.2 & 52.3 & 73.8 & 61.3 & 56.6 & 62.7 & 55.4 & 50.7 \\

\bottomrule
\end{tabular}
\end{adjustbox}
\label{tab:weight_comparison}
\vspace{-2mm}
\end{table*}

\end{document}

%% file: sections/01_intro.tex

\begin{abstract}
High-fidelity generative models have narrowed the perceptual gap between synthetic and real images, posing serious threats to media security. Most existing AI-generated image (AIGI) detectors rely on artifact-based classification and struggle to generalize to evolving generative traces. In contrast, human judgment relies on stable real-world regularities, with deviations from the human cognitive manifold serving as a more generalizable signal of forgery. Motivated by this insight, we reformulate AIGI detection as a Reference-Comparison problem that verifies consistency with the real-image manifold rather than fitting specific forgery cues.
We propose \textbf{MIRROR} (\textbf{M}anifold \textbf{I}deal \textbf{R}eference \textbf{R}econstruct\textbf{OR}), a framework that explicitly encodes reality priors using a learnable discrete memory bank. MIRROR projects an input into a manifold-consistent ideal reference via sparse linear combination, and uses the resulting residuals as robust detection signals. To evaluate whether detectors reach the “superhuman crossover” required to replace human experts, we introduce the \textbf{Human-AIGI} benchmark, featuring a psychophysically curated human-imperceptible subset. Across 14 benchmarks, MIRROR consistently outperforms prior methods, achieving gains of 2.1\% on six standard benchmarks and 8.1\% on seven in-the-wild benchmarks. On Human-AIGI, MIRROR reaches 89.6\% accuracy across 27 generators, surpassing both lay users and visual experts, and further approaching the human perceptual limit as pretrained backbones scale. The code is publicly available at: https://github.com/349793927/MIRROR
\end{abstract}


\section{Introduction}
\label{intro}

\begin{figure}[t]
    \centering
    \includegraphics[scale=0.435]{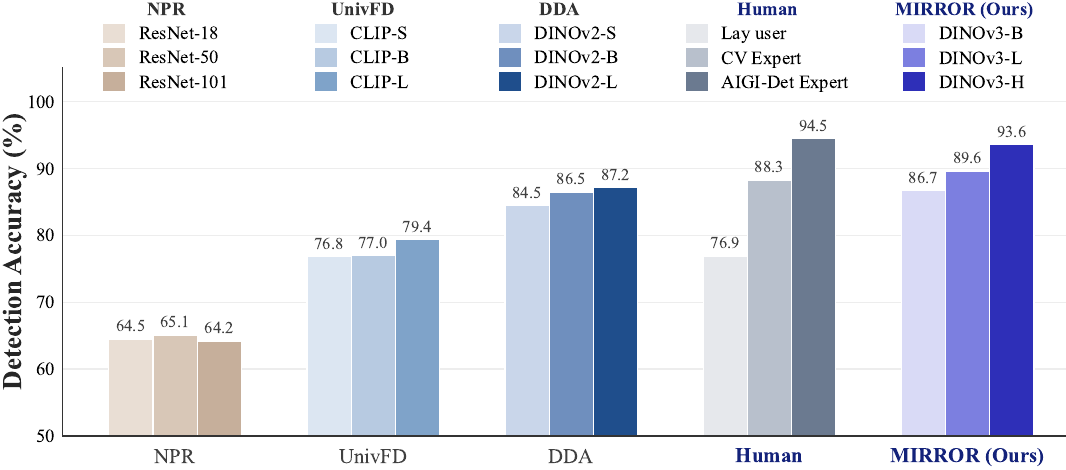}
\caption{\textbf{Scalability comparison of AIGI detection between detectors and humans on Human-AIGI.} While existing methods (NPR~\cite{tan2024rethinking}, UnivFD~\cite{ojha2023towards}, and DDA~\cite{chen2025dual}) exhibit saturated performance as model size increases, humans demonstrate strong prior-driven scalability. In contrast, \textbf{MIRROR} exhibits similar human-like behavior, achieving sustained performance gains as model capacity scales from Base to Huge.}
    \label{fig:logit_stability}
    \vspace{-5mm}
\end{figure}


\epigraph{\textit{``Perception is a process of hypothesis testing.''}}{--- Richard L. Gregory, 1980}
\vspace{-3mm}
High-fidelity generative models have largely closed the perceptual gap between synthetic and real images, making generated content often indistinguishable from authentic photographs~\cite{goodfellow2014generative,ho2020denoising,rombach2022high,betker2023dalle3}. This trend threatens the reliability of online media, forensic analysis, and downstream vision systems, \textit{highlighting the urgent need for detection frameworks that can surpass human visual perception.} 

Most AIGI detectors treat real-versus-fake recognition as an isolated binary classification problem: they assume that the presence of artifacts implies a generated image, and their absence implies a real image, performing detection by estimating the probability of artifacts~\cite{wang2020cnn,frank2020leveraging,chai2020makes,qian2020thinking}. This strategy works well under ideal conditions, where artifacts are salient or unperturbed. However, when confronted with unknown generators, performance degrades substantially, indicating severe generalization limitations.
This raises a central question: \textit{as model capacity continues to scale and pretrained priors become increasingly powerful, can existing detection paradigms still achieve sustained improvements in generalization?}

We systematically compare the detection performance of existing detectors and humans on previously unseen generated images and observe a counterintuitive phenomenon: the performance gains from pretrained prior knowledge exhibit only limited scaling behavior (see Fig.~\ref{fig:logit_stability}). Although methods based on pretrained models~\cite{ojha2023towards, chen2025dual} significantly outperform those without pretraining~\cite{tan2024rethinking}, their performance improvements quickly saturate as the backbone model size increases.

We argue that this phenomenon does not stem from deficiencies in pretrained knowledge itself. Instead, it arises from intrinsic limitations in how current detection paradigms leverage such knowledge. In existing approaches, pretrained models are typically treated as generic feature extractors that fit artifact cues left by generative models (Fig.~\ref{fig:intro}, left). They are rarely used to explicitly model stable regularities of the real world. As a result, detectors approximate a decision boundary of the forged distribution, which continuously evolves as generative algorithms advance. Because this forged distribution is inherently non-stationary, artifact-driven methods are prone to overfitting and cross-domain failure. This also explains why scaling pretrained backbones does not yield sustained improvements in generalization. 

In contrast, humans exhibit substantially stronger scalability in AIGI detection. Visual experts with extensive exposure to real images significantly outperform non-experts. Human detection accuracy further improves when reference images are provided. We argue that this advantage does not arise from memorizing specific generative artifacts. Instead, humans treat prior knowledge as robust, method-agnostic regularities of real-world images. As illustrated in Fig.~\ref{fig:intro} (Middle), these priors are used to construct an internal ideal reference and detect deviations by comparing the observed image against this reference~\cite{cox2014neural, costa2025anterior}. Because this process is anchored to the stable real-image distribution rather than the non-stationary forged distribution, human judgments exhibit stronger cross-generator generalization and scalability (see Sec.~\ref{sec:motivation}).

Motivated by these findings, we propose \textbf{MIRROR} (\textbf{M}anifold \textbf{I}deal \textbf{R}eference \textbf{R}econstruct\textbf{OR}), the first framework to formulate forgery detection as a \textbf{Reference-Comparison} process grounded in priors. Under the manifold hypothesis~\cite{bengio2013representation,cayton2005algorithms}, real images lie on a low-dimensional manifold embedded in high-dimensional pixel space. As illustrated in Fig.~\ref{fig:intro}(Right), MIRROR encodes this real-manifold prior using a learnable discrete memory bank composed of orthogonal prototype vectors, which capture stable texture and semantic patterns from real images. Given an input image, MIRROR retrieves a small set of prototypes and constructs a manifold-consistent reconstruction via sparse linear combination, serving as the \textbf{Ideal Reference}. The detection signal is derived from the deviation between the observed input and this ideal Reference. Since real images conform to the natural statistics encoded in the memory, they align closely with their Reference; in contrast, generated images exhibit subtle physical inconsistencies (e.g., in illumination or geometry) that cannot be explained by the reality-based memory, resulting in significant deviations.

Existing benchmarks primarily emphasize the number of generators or semantic diversity, yet lack a definitive standard to verify when machine detection surpasses human capabilities. To measure this gap and identify when AI can reliably replace human expert reviewers, we introduce the \textbf{Human-AIGI} Benchmark. Through psychophysical experiments with 50 participants, we record accuracy, confidence, and response latency, curating a Human-Imperceptible subset comprising samples that deceive humans or induce hesitation. This benchmark serves as a quantitative yardstick for determining the ``Superhuman Crossover'', when algorithmic performance exceeds biological perception limits.

Comprehensive experiments across 14 benchmarks demonstrate that MIRROR consistently exhibits superior generalization and robustness. On six standard benchmarks, MIRROR improves the average accuracy over state-of-the-art methods by 2.1\%, while achieving a 8.1\% average accuracy gain on seven in-the-wild benchmarks. On the human-imperceptible subset of \textbf{Human-AIGI}, MIRROR attains an accuracy of 89.5\%, outperforming prior methods by 2.6\%. Moreover, as shown in Fig.\ref{fig:logit_stability}, as the pretrained backbone scales up, MIRROR’s accuracy further approaches the human perceptual limit (i.e., AIGI detection experts). These results indicate that MIRROR can serve as a practical and reliable assistive tool for AIGI detection.

Our main contributions are summarized as follows:
\begin{itemize}
\item We identify a key limitation of existing detectors: chasing evolving generative artifacts leads to poor scalability. Inspired by human vision, we reformulate AIGI detection as a \textbf{Reference-Comparison} process that anchors decisions on invariant reality priors by comparing inputs against an internal reference.

\item We propose MIRROR, a novel framework that encodes real-world regularities using a learnable memory bank of orthogonal prototypes. MIRROR constructs an \textbf{Ideal Reference} by sparsely projecting input features onto the real-image manifold and identifies forgeries through reconstruction residuals, enabling robust, generator-agnostic detection.

\item We introduce the \textbf{Human-AIGI} Benchmark, including a Human-Imperceptible subset curated through psychophysical experiments, to quantify when detectors surpass human perception. Evaluations across 14 benchmarks show that MIRROR achieves strong generalization on human-imperceptible samples, marking the ``Superhuman Crossover''.
\end{itemize}

%% file: sections/02_related_work.tex
\section{Related Work} \label{sec:related}

\subsection{AI-Generated Image Detection}

AI-generated image (AIGI) detection has been extensively studied, with a central focus on generalization to unseen generators~\citep{wang2020cnn, ojha2023towards}. Early approaches primarily relied on low-level artifacts, exploiting generator-specific fingerprints in the spatial or spectral domain~\citep{durall2020watch, frank2020leveraging}. CNNSpot~\citep{wang2020cnn} demonstrated cross-GAN transferability, and subsequent works explored frequency irregularities~\citep{qian2020thinking, tan2024frequency}, upsampling traces~\citep{tan2024rethinking}. However, such texture-level cues are fragile under generator evolution and real-world post-processing~\citep{liu2025beyond}.

To improve robustness, recent methods increasingly leverage pre-trained foundation models. UnivFD~\citep{ojha2023towards} combines CLIP~\citep{radford2021learning} features with lightweight classifiers, while FatFormer~\citep{liu2024forgery} and AIDE~\citep{yan2024sanity} adapt CLIP representations via task-specific modules. Effort~\citep{yan2024orthogonal} further exploits pretraining by disentangling semantic and forgery-related subspaces.

Another line of work attributes performance degradation to data bias, and mitigates shortcut learning through reconstruction or alignment between real and synthetic data. Representative methods include FakeInversion~\citep{cazenavette2024fakeinversion}, SemGIR~\citep{yu2024semgir}, DRCT~\citep{chen2024drct}, B-Free~\citep{guillaro2025bias}, AlignedForensics~\citep{rajan2024aligned}, and Dual Data Alignment~\citep{chen2025dual}. Despite these advances, most existing detectors remain black-box discriminators and lack explicit, hierarchical priors and verification mechanisms comparable to human perception.

\subsection{Detection Benchmarks}
Early AIGI detection benchmarks mainly target GAN imagery, exemplified by the CNNSpot dataset~\citep{wang2020cnn}. Later datasets substantially broaden generator coverage: GenImage~\citep{zhu2023genimage} and AIGCDetect~\citep{zhong2023patchcraft} provide large-scale samples from diverse models, while DRCT-2M~\citep{chen2024drct} offers million-level data tailored for diffusion reconstruction. Evaluation protocols have also been strengthened, including UniversalFakeDetect~\citep{ojha2023towards} for cross-model testing with CLIP backbones, Synthbuster~\citep{bammey2023synthbuster} for systematic diffusion evaluation, and EvalGEN~\citep{chen2025dual} for emerging autoregressive generators. Despite their scale, many of these benchmarks rely on randomly synthesized images with simple prompts, which may underrepresent the complexity of human-targeted forgeries.

To better reflect real-world conditions, several in-the-wild benchmarks collect web images or simulate realistic degradation. Chameleon~\citep{yan2024sanity} curates high-resolution samples that are challenging even for humans, exposing detector generalization issues. SynthWildx~\citep{zhong2023patchcraft}, WildRF~\citep{cavia2024real}, and AIGIBench~\citep{li2025artificial} further expand web diversity, while RRDataset~\citep{li2025bridging} and RealChain~\citep{liu2025beyond} model post-processing and propagation to stress-test robustness. 
Nevertheless, most existing benchmarks emphasize generator diversity or semantic coverage, but do not quantify where human perception starts to fail. As a result, they cannot directly measure whether a detector truly improves upon human judgment on the most deceptive samples.

%% file: sections/03_method.tex
\begin{figure*}[t]
    \centering
    \includegraphics[scale=0.148]{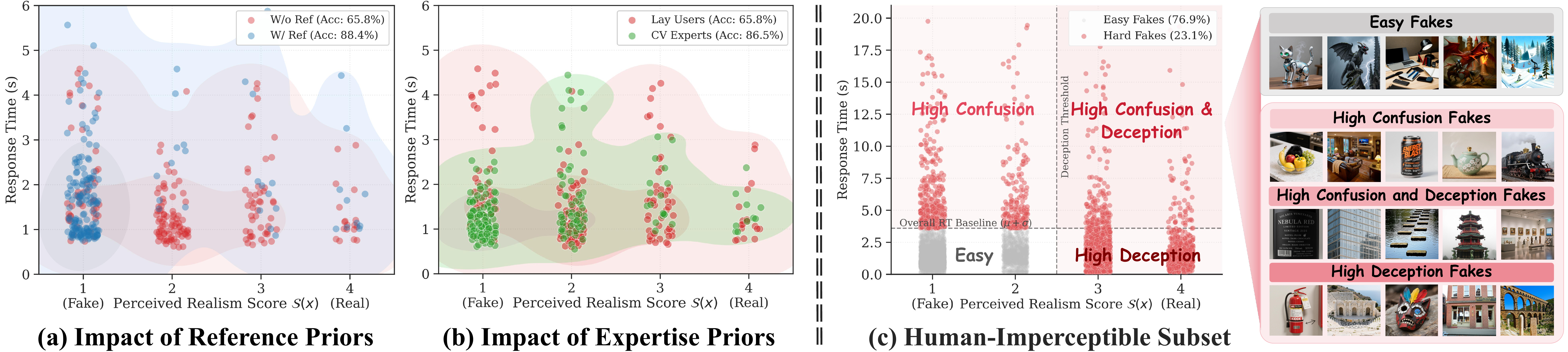} 
    \caption{
    \textbf{Psychophysical Analysis on the Human-AIGI Benchmark.}
    Results are obtained from controlled psychophysical experiments on AI-generated images from 27 generators.
    \textbf{(a) Reference Priors.} Providing explicit reference images shifts hard samples (red) from high uncertainty with long response times to a higher-confidence regime (blue).
    \textbf{(b) Expertise Priors.} CV experts (green) consistently outperform lay users (red), reflecting the benefit of stronger internal perceptual priors.
    \textbf{(c) Human-Imperceptible Subset.} The \emph{Hard} subset (red) includes samples with high deception or confusion, measured by confidence and response time; unlike \emph{Easy Fakes} (gray), these images preserve high visual fidelity and define a rigorous benchmark.
    }
    \label{fig:human_aigi}
    \vspace{-3mm}
\end{figure*}

\section{Motivation} 
\label{sec:motivation}
Our motivation stems from a fundamental difference in decision paradigms between machine and human vision. Existing detectors typically formulate AIGI detection as artifact classification, which overfits generator-specific traces and degrades as generative models evolve. In contrast, humans adopt a Reference–Comparison strategy, anchoring judgments in long-term understanding of invariant physical regularities rather than artifact memorization. This paradigm aligns with the cognitive theory that perception operates as hypothesis testing. Neuroscience studies show that the visual cortex functions as an active inference system, generating internal predictions as references and continuously comparing them with sensory input; mismatches between observation and reference elicit strong neural responses~\cite{costa2025anterior,cox2014neural}. This mechanism supports robust perception in unfamiliar environments.

Our psychophysical experiments further validate the effectiveness of this underlying mechanism for detecting generated images. Results show that even when lay user have never been systematically exposed to large-scale generated imagery, providing explicit real-image references significantly improves their detection accuracy (see Fig.~\ref{fig:human_aigi}(a)). Moreover, computer vision experts (though not forensic experts), relying solely on priors acquired through prolonged exposure to real images, substantially outperform lay user (see Fig.~\ref{fig:human_aigi}(b)). These findings indicate that human superiority arises not from artifact memorization, but from the choice of a stable decision anchor. By grounding detection in the stable real-image distribution, generated image detection becomes a problem of consistency verification, yielding stronger cross-generator generalization than conventional artifact-driven methods.

Inspired by this insight, we introduce MIRROR, which explicitly models the real-image manifold and constructs a manifold-consistent Ideal Reference for each input. Detection is then performed by measuring the Comparison Residual between the observation and its reference, resulting in a robust, generator-agnostic framework.

\begin{figure*}[t!]
	\centering
	\includegraphics[scale=0.61]{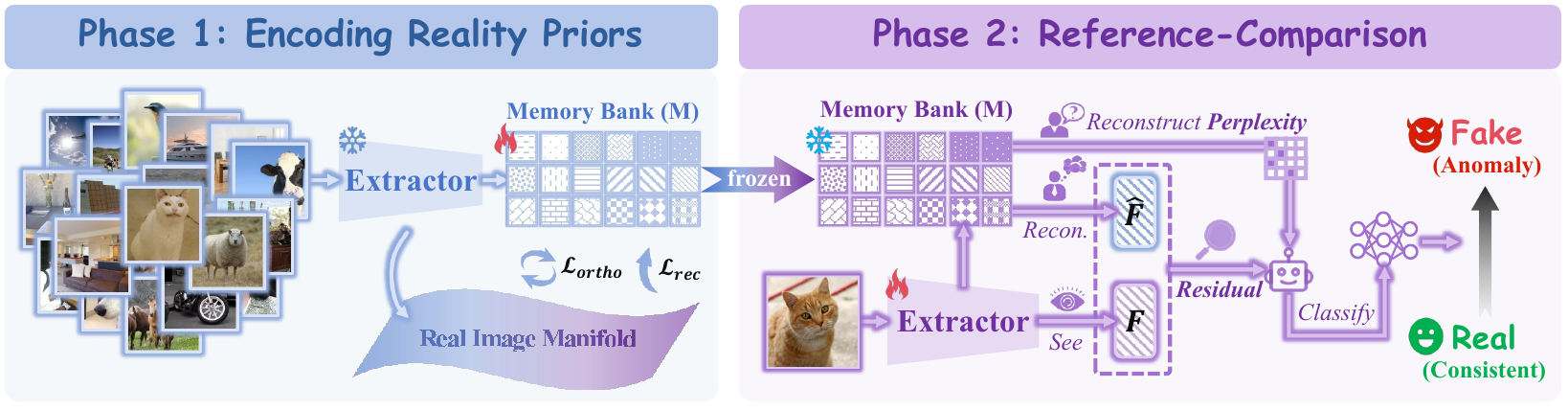}
\caption{\textbf{Architecture of MIRROR.} The framework operates in two distinct phases. In Phase 1, a frozen DINO encoder extracts patch-level features from real images to encode Reality Priors within an orthogonal memory bank that approximates the real-image manifold. In Phase 2, the learned priors are frozen. Given an input image, MIRROR constructs a manifold-consistent Ideal Reference through sparse projection and identifies forgeries via Reference-Comparison. This process utilizes both the reconstruct perplexity and the comparison residual to evaluate whether the input deviates from the learned reality manifold.}
	\label{fig_method}
    \vspace{-3mm}
\end{figure*}

\section{Methodology}
\label{sec:method}
Our MIRROR formalizes the human reference–comparison paradigm into a learnable two-stage framework (see Fig.\ref{fig_method}): first encoding stable Reality Priors from real images to construct a Manifold-consistent Ideal Reference, and then performing robust detection by analyzing the Comparison Residual between the input and its reference.

\subsection{Phase 1: Encoding Reality Priors}
\label{sec:phase1}

Phase~1 aims to learn a latent feature space that approximates the real-image manifold $\mathcal{M}_{real}$. To capture the local geometry and fine-grained textures essential for physical consistency, we utilize a frozen DINO\cite{simeoni2025dinov3} encoder $\mathcal{E}$ to extract patch-level features:
\begin{equation}
F = \mathcal{E}(I) \in \mathbb{R}^{N \times D},
\end{equation}
where $N$ denotes the number of patches and $D$ is the feature dimension.
We encode reality priors into a discrete memory bank $\mathbf{M} \in \mathbb{R}^{K \times D}$ containing $K$ orthogonal prototypes. Each prototype is optimized to represent a frequent, stable pattern found in the real-world distribution. To ensure the generated \textbf{Ideal Reference} remains faithful to $\mathcal{M}_{real}$, we project the input features $F$ onto the subspace spanned by $\mathbf{M}$ using a cross-attention mechanism with a Top-$k$ sparsity constraint:
\begin{equation}
\mathbf{A} = \text{Softmax} \left( \text{Top-}k \left( \frac{QK^\top}{\sqrt{D}} \right) \right), \quad \hat{F} = \mathbf{A}V,
\label{eq:attn}
\end{equation}
where $Q$ is derived from $F$, and $K, V$ are derived from $\mathbf{M}$. The Top-$k$ operation ensures that $\hat{F}$ is a sparse linear combination of real-world prototypes, effectively filtering out patterns that do not belong to the real-image manifold. 

The training of Phase~1 is conducted exclusively on real images using the following objective:
\begin{equation}
\mathcal{L}_{Phase1} = \|F - \hat{F}\|_2^2 + \lambda \|\mathbf{M}\mathbf{M}^\top - \mathbf{I}\|_F,
\end{equation}
where the first term minimizes the reconstruction error to ensure the \textbf{Ideal Reference} $\hat{F}$ accurately reflects the input's real components, while the second term enforces orthogonality among prototypes to maximize the diversity and non-redundancy of the encoded reality priors.

\subsection{Phase 2: Reference-Comparison-Based Detection}
\label{sec:phase2}

In Phase~2, we freeze the memory bank $\mathbf{M}$ and the projection module defined in Eq.~\ref{eq:attn} to ensure that the learned Reality Priors remain independent of the forgery distribution. We perform detection by evaluating the consistency between the input feature $F$ and its \textbf{Ideal Reference} $\hat{F}$ through two complementary signals. Specifically, the \textbf{Reconstruct Perplexity} characterizes the uncertainty in prototype retrieval when an input lacks typical real-world patterns, while the \textbf{Comparison Residual} captures the irreconstructible details that fall outside the real-image manifold. We derive the perplexity from the attention matrix $\mathbf{A}$ (Eq.\ref{eq:attn}) using the maximum attention score $s_{max}$ and entropy $s_{ent}$, and calculate the residual as $\Delta F = F - \hat{F}$ to quantify manifold deviations. These two signals are projected into evidence vectors $V_{per}$ and $V_{res}$ respectively:
\begin{equation}
V_{per} = \text{MLP}_{per}([s_{max}, s_{ent}]), \quad V_{res} = \text{Linear}(\Delta F).
\end{equation}
The final forgery probability $y_{pred}$ is predicted by a classification head $\text{MLP}_{\mathcal{C}}$ acting on the concatenated evidence:
\begin{equation}
y_{pred} = \text{MLP}_{\mathcal{C}} \left( \text{Concat}[V_{per}, V_{res}] \right).
\end{equation}
The framework is optimized by a binary cross-entropy loss:
\begin{equation}
\mathcal{L}_{Phase2} = - y \log y_{pred} - (1-y) \log(1-y_{pred}),
\end{equation}
where $y \in \{0, 1\}$ is the ground-truth label. By anchoring the decision on the Reference-Comparison of these dual signals, MIRROR identifies forgeries as instances that the real-world manifold cannot sparsely reconstruct or accurately explain.

\section{The Human-AIGI Benchmark} 
\label{sec:human_benchmark}

\subsection{Psychophysical Experimental Design}

We conduct large-scale psychophysical experiments to probe the limits of human perception in AIGI detection. We recruit 50 participants, including lay users, computer vision experts, and AIGI detection experts, and evaluate them on a dataset $\mathcal{D}_{orig.}$ comprising over 30,000 images synthesized by 27 generators from T2I-COREBENCH~\cite{li2025easier}. Participants perform a controlled binary classification task, judging whether each image is real or AI-generated without external assistance.
Beyond accuracy, we record two key psychophysical variables for each trial: perceived realism (confidence score $S$) and response time ($RT$). These measurements capture not only decision outcomes but also the cognitive effort and uncertainty involved in distinguishing synthetic images from real ones. Lay users aggregated results are shown in Fig.~\ref{fig:human_aigi} (c), providing insight into how human judgments rely on stable real-world priors rather than generator-specific artifacts.

\subsection{Definition of the Human-Imperceptible Subset}

Based on these observations, we define a \emph{human-imperceptible subset} consisting of generated samples that challenge human perception. Such samples either deceive observers into confidently judging them as real or induce significant hesitation during decision-making.
Formally, following psychophysical principles, we define the human-imperceptible subset $\mathcal{D}_{hard}$ as:
\begin{equation}
\mathcal{D}_{hard} = \{ x \in \mathcal{X}_{gen} \mid S(x) \ge \tau_{real} \lor RT(x) > \mu_{rt} + \sigma_{rt} \},
\end{equation}
where $\mathcal{X}_{gen}$ denotes AI-generated images, $S(x) \in \{1,2,3,4\}$ is the perceived realism score, $\tau_{real}$ is the deception threshold, and $\mu_{rt}, \sigma_{rt}$ are the mean and standard deviation of response times across the benchmark.
By focusing on samples that appear highly realistic or induce pronounced uncertainty, this subset provides a stringent testbed for assessing whether detection algorithms can reliably identify generative threats that evade human perceptual scrutiny.
By jointly evaluating performance on $\mathcal{D}_{orig.}$ and $\mathcal{D}_{hard}$, we can assess whether a detector has reached the milestone of reliably replacing human expert review.

%% file: sections/04_experiments.tex
\vspace{-2mm}
\begin{table*}[t!]
\centering
\caption{Performance comparison on \textbf{Standard Benchmarks}. We report Balanced Accuracy (B.Acc), JPEG Robustness (J.Rob), and the Resize Robustness (R.Rob) to comprehensively evaluate the detector's generalization and robustness performance. The best and second-best results are highlighted in \textbf{bold} and \underline{underline}, respectively. DRCT~\cite{chen2024drct}, Aligned~\cite{rajan2024aligned}, B-Free~\cite{guillaro2025bias}, and DDA~\cite{chen2025dual} are evaluated using their officially released weights. Other methods based on architectural innovations are evaluated using the same training set as ours, and evaluation results using their official weights are additionally reported in the appendix.}
\vspace{-2mm}
\fontsize{9pt}{11pt}\selectfont
\setlength{\tabcolsep}{5pt}
\renewcommand{\arraystretch}{1}
\begin{adjustbox}{max width=\textwidth}
\begin{tabular}{l | c c c | c c c | c c c | c c c | c c c | c c c | c c c}
\toprule
\multirow{2}{*}{\textbf{Method}} & 
\multicolumn{3}{c|}{\textbf{AIGCDetect}} & 
\multicolumn{3}{c|}{\textbf{GenImage}} & 
\multicolumn{3}{c|}{\textbf{DRCT-2M}} & 
\multicolumn{3}{c|}{\textbf{UnivFakeDetect}} & 
\multicolumn{3}{c|}{\textbf{Synthbuster}} & 
\multicolumn{3}{c|}{\textbf{EvalGEN}} & 
\multicolumn{3}{c}{\textbf{Average}} \\
\cmidrule(lr){2-4} \cmidrule(lr){5-7} \cmidrule(lr){8-10} \cmidrule(lr){11-13} \cmidrule(lr){14-16} \cmidrule(lr){17-19} \cmidrule(lr){20-22}
  & B.Acc & J.Rob & R.Rob & B.Acc & J.Rob & R.Rob & B.Acc & J.Rob & R.Rob & B.Acc & J.Rob & R.Rob & B.Acc & J.Rob & R.Rob & B.Acc & J.Rob & R.Rob & B.Acc & J.Rob & R.Rob \\
\midrule
NPR \cite{tan2024rethinking}       & 63.9 & 64.0 & 48.2 & 73.7 & 73.7 & 46.6 & 59.5 & 58.3 & 37.4 & 51.1 & 51.1 & 49.2 & 64.6 & 64.0 & 34.4 & 66.1 & 65.7 & 64.2 & 63.2 & 62.8 & 46.7 \\
UnivFD \cite{ojha2023towards}      & 56.5 & 50.1 & 48.7 & 62.5 & 53.6 & 50.8 & 69.5 & 57.3 & 52.5 & 48.3 & 51.1 & 43.0 & 65.8 & 59.2 & 52.3 & 73.8 & 61.3 & 56.6 & 62.7 & 55.4 & 50.7 \\
FatFormer \cite{liu2024forgery}     & 82.1 & 71.5 & 80.0 & 71.5 & 58.4 & 68.4 & 53.9 & 52.1 & 52.4 & 86.7 & 68.9 & 85.2 & 69.2 & 57.1 & 65.6 & 56.7 & 53.4 & 57.5 & 70.0 & 60.2 & 68.2 \\
SAFE \cite{li2024improving}         & 48.4 & 57.8 & 49.4 & 47.7 & 64.4 & 49.4 & 50.3 & 62.4 & 49.2 & 49.7 & 48.5 & 49.8 & 51.6 & 53.8 & 49.9 & 50.2 & 55.4 & 49.1 & 49.7 & 57.1 & 49.5 \\
C2P-CLIP \cite{tan2025c2p}          & 80.0 & 65.7 & 77.6 & 71.1 & 55.6 & 67.8 & 54.4 & 54.0 & 55.7 & 87.3 & 68.8 & 85.2 & 69.4 & 57.3 & 66.4 & 56.8 & 52.1 & 65.8 & 69.8 & 58.9 & 69.8 \\
AIDE \cite{yan2024sanity}          & 84.0 & 53.4 & 77.0 & 88.6 & 55.3 & 76.3 & 59.2 & 54.6 & 72.5 & 77.9 & 51.5 & 70.8 & 75.4 & 76.8 & 62.5 & 59.5 & 58.8 & 76.4 & 74.1 &  58.4 & 72.6 \\
DRCT \cite{chen2024drct}           & 67.1 & 67.1 & 67.5 & 78.8 & 78.5 & 80.3 & 97.0 & 93.0 & 98.4 & 66.8 & 69.5 & 69.7 & 81.0 & 80.6 & 83.3 & 81.9 & 85.5 & 82.4 & 78.8 & 79.0 & 80.3 \\
Aligned \cite{rajan2024aligned}    & 55.7 & 54.5 & 55.3 & 57.5 & 55.9 & 57.0 & 54.9 & 54.7 & 54.3 & 60.5 & 59.7 & 61.0 & 54.5 & 53.0 & 53.8 & 65.8 & 65.2 & 67.5 & 58.2 & 57.2 & 58.2 \\
B-Free \cite{guillaro2025bias}   & 84.7 & 85.0 & 82.7 & 89.6 & 90.2 & 87.9 & 99.2 & 99.0 & 98.2 & 87.8 & 87.0 & 86.2 & 95.7 & 95.9 & 94.8 & 94.6 & 93.7 & 92.5 & {91.9} & {91.8} & {90.4} \\
DDA \cite{chen2025dual}          & 81.5 & 81.8 & 79.1 & 88.9 & 88.7 & 88.1 & 97.0 & 97.7 & 95.6 & 69.3 & 69.3 & 66.7 & 96.5 & 95.6 & 94.6 & 96.6 & 90.1 & 93.9 & 88.3 & 87.2 & 86.3 \\
\midrule
\rowcolor{blue!5}
\textbf{MIRROR (DINOv2)}  &90.5&90.0&88.6&91.3&94.1&93.7&92.8&88.9&94.2&84.6& 85.8& 85.0&97.0& 96.4 & 97.6 & 98.1& 97.1 & 96.6 &\underline{92.4}{\color{red}\scriptsize $\uparrow$0.5}& \underline{92.1}{\color{red}\scriptsize $\uparrow$0.3}& \underline{92.6}{\color{red}\scriptsize $\uparrow$2.2} \\
\rowcolor{blue!5}
\textbf{MIRROR (DINOv3)}  & 91.7 & 91.5 & 91.5 & 94.2 & 96.7 & 94.9 & 93.0 & 90.4 & 93.4 & 88.2 & 88.3 & 87.5 & 98.1 & 97.5 & 97.6 & 99.0 & 98.5 & 98.6 & \textbf{94.0}{\color{red}\scriptsize $\uparrow$2.1} & \textbf{93.8}{\color{red}\scriptsize $\uparrow$2.0} & \textbf{93.9}{\color{red}\scriptsize $\uparrow$3.5} \\
\bottomrule
\end{tabular}
\end{adjustbox}
\label{tab:standard_benchmarks}
\end{table*}

\begin{table*}[t!]
\centering
\caption{Performance comparison on \textbf{In-the-wild Benchmarks}. These datasets have undergone low-level degradation caused by network propagation. The best and second-best results in the summary columns are highlighted in \textbf{bold} and \underline{underline}, respectively.}
\vspace{-2mm}
\fontsize{9pt}{11pt}\selectfont
\setlength{\tabcolsep}{5pt}
\renewcommand{\arraystretch}{1}
\begin{adjustbox}{max width=\textwidth}
\begin{tabular}{l c | c c c | c c c | c c | c c c c c | c | c | c}
\toprule 
\multirow{2}{*}{\textbf{Method}} & 
\multirow{2}{*}{\textbf{\makecell{Cham-\\eleon}}} & 
\multicolumn{3}{c|}{\textbf{SynthWildx}} & 
\multicolumn{3}{c|}{\textbf{WildRF}} & 
\multicolumn{2}{c|}{\textbf{AIGIBench}} & 
\multicolumn{5}{c|}{\textbf{CO-SPY-Bench/in-the-wild}} & 
\multirow{2}{*}{\textbf{\makecell{RR-\\Dataset}}} & 
\multirow{2}{*}{\textbf{\makecell{BFree-\\Online}}} & 
\multirow{2}{*}{\textbf{\makecell{Avg\\B.Acc}}}
\\
\cmidrule(lr){3-5} \cmidrule(lr){6-8} \cmidrule(lr){9-10} \cmidrule(lr){11-15}

 & & DALLE3 & Firefly & Midj. & FB & Reddit & Twitter & SocRF & ComAI & Civitai & DALLE3 & instavibe.ai & Lexica & Midj.v6 & & & \\
\midrule
NPR \cite{tan2024rethinking}     & 55.2 & 62.9 & 53.8 & 63.0 & 53.8 & 53.7 & 57.7 & 55.8 & 52.0 & 51.3 & 74.4 & 6.0 & 3.9 & 80.7 & 48.3 & 40.5 & 50.8 \\
UnivFD \cite{ojha2023towards}    & 39.5 & 52.5 & 52.4 & 50.7 & 54.1 & 55.1 & 66.5 & 51.6 & 45.4 & 89.4 & 76.6 & 78.3 & 63.1 & 84.1 & 51.1 & 57.2 & 60.5 \\
FatFormer \cite{liu2024forgery}  & 57.8 & 52.6 & 56.6 & 50.0 & 52.5 & 65.4 & 39.7 & 55.7 & 50.4 & 0.4 & 2.2 & 0.0 & 0.3 & 15.2 & 50.4 & 32.7 & 36.4 \\
SAFE \cite{li2024improving}      & 56.8 & 48.8 & 46.6 & 48.7 & 49.7 & 49.2 & 34.7 & 49.2 & 49.3 & 0.1 & 0.9 & 0.5 & 0.3 & 2.4 & 49.3 & 32.6 & 29.3 \\
C2P-CLIP \cite{tan2025c2p}       & 57.6 & 49.6 & 57.9 & 49.6 & 51.9 & 67.6 & 40.4 & 58.1 & 50.4 & 0.0 & 4.5 & 0.2 & 0.2 & 23.2 & 50.0 & 32.7 & 37.1 \\
AIDE \cite{yan2024sanity}        & 65.7 & 66.4 & 48.2 & 66.4 & 61.6 & 66.3 & 52.5 & 59.2 & 62.2 & 15.8 & 34.0 & 7.2 & 19.5 & 64.5 & 57.6 & 52.1 & 50.0 \\
DRCT \cite{chen2024drct}         & 79.8 & 85.9 & 58.9 & 90.5 & 90.3 & 66.8 & 79.6 & 71.3 & 84.6 & 99.1 & 82.1 & 22.0 & 55.8 & 94.8 & 58.2 & 77.1 & 81.1 \\
Aligned \cite{rajan2024aligned}  & 61.3 & 49.6 & 53.9 & 52.4 & 48.4 & 54.0 & 40.6 & 51.0 & 60.2 & 7.5 & 8.4 & 6.2 & 11.5 & 9.7 & 47.7 & 38.1 & 37.5 \\
B-Free \cite{guillaro2025bias}   & 78.3 & 96.1 & 92.3 & 95.3 & 95.6 & 85.5 & 96.7 & 84.9 & 79.7 & 98.8 & 97.4 & 5.9 & 53.6 & 98.6 & 69.5 & 87.1 & 82.2 \\
DDA \cite{chen2025dual}          & 83.5 & 91.1 & 84.7 & 91.6 & 85.3 & 82.5 & 89.3 & 79.9 & 88.9 & 99.7 & 94.6 & 35.4 & 73.7 & 97.9 & 70.3 & 81.2 & 
83.1 \\
\midrule
\rowcolor{blue!5}
\textbf{MIRROR (DINOv2)} & 85.4&91.2&86.1	&89.5&	93.8&	90.2&	92.7&	82.3	&88.9	&94.6	&93.4	&72.5	&91.1	&85.6	&76.8&	84.3& \underline{87.4}{\color{red}\scriptsize $\uparrow$4.3}  \\

\rowcolor{blue!5}
\textbf{MIRROR (DINOv3)} & 90.7 & 95.9 & 88.4 & 94.9 & 97.1 & 96.6 & 96.4 & 87.6 & 93.4 & 99.2 & 98.1 & 75.0 & 94.2 & 89.8 & 78.9 & 83.0 & \textbf{91.2}{\color{red}\scriptsize $\uparrow$8.1} \\
\bottomrule
\end{tabular}
\end{adjustbox}
\label{tab:wild}
\vspace{-3mm}
\end{table*}

\begin{table*}[t]
    \centering
     \caption{Performance comparison on the \textbf{Human-AIGI Benchmark}. We report detection accuracy (\%) on the original test set (\textbf{Orig.}) and the curated \textbf{Human-Imperceptible} subset (\textbf{Hard}).}
    \vspace{-2mm}
    \label{tab:main_comparison}
    \resizebox{\linewidth}{!}{
        \begin{tabular}{l cc cc cc cc cc cc cc cc cc cc cc}
            \toprule
            \multirow{2}{*}{\textbf{Generator}} & 
            \multicolumn{2}{c}{\textbf{NPR}} & 
            \multicolumn{2}{c}{\textbf{UnivFD}} & 
            \multicolumn{2}{c}{\textbf{C2P}} & 
            \multicolumn{2}{c}{\textbf{FatFormer}} & 
            \multicolumn{2}{c}{\textbf{SAFE}} & 
            \multicolumn{2}{c}{\textbf{AIDE}} & 
            \multicolumn{2}{c}{\textbf{DRCT}} & 
            \multicolumn{2}{c}{\textbf{Aligned}} & 
            \multicolumn{2}{c}{\textbf{B-Free}} & 
            \multicolumn{2}{c}{\textbf{DDA}} & 
            \multicolumn{2}{c}{\cellcolor{blue!5}\textbf{MIRROR (Ours)}} \\
            
            \cmidrule(lr){2-3} \cmidrule(lr){4-5} \cmidrule(lr){6-7} \cmidrule(lr){8-9} \cmidrule(lr){10-11} 
            \cmidrule(lr){12-13} \cmidrule(lr){14-15} \cmidrule(lr){16-17} \cmidrule(lr){18-19} \cmidrule(lr){20-21} \cmidrule(lr){22-23}
            
            & Orig. & Hard & Orig. & Hard & Orig. & Hard & Orig. & Hard & Orig. & Hard & Orig. & Hard & Orig. & Hard & Orig. & Hard & Orig. & Hard & Orig. & Hard & \cellcolor{blue!5}Orig. & \cellcolor{blue!5}Hard \\
            \midrule
            
            \multicolumn{23}{c}{\textit{Diffusion Models}} \\
            \midrule
            SD-3-Medium       & 61.2 & 52.4 & 80.3 & 76.2 & 1.2 & 0.0 & 0.8 & 0.0 & 32.0 & 42.9 & 34.4 & 28.6 & 66.2 & 76.2 & 12.2 & 0.0 & 93.9 & 100.0 & 94.5 & 95.2 & \cellcolor{blue!5}91.1 & \cellcolor{blue!5}100.0 \\
            SD-3.5-Medium    & 68.8 & 90.6 & 87.0 & 84.4 & 2.4 & 0.0 & 2.2 & 0.0 & 28.6 & 21.9 & 12.4 & 6.3 & 52.2 & 46.9 & 16.8 & 9.4 & 97.5 & 93.8 & 93.9 & 93.8 & \cellcolor{blue!5}92.3 & \cellcolor{blue!5}93.7 \\
            SD-3.5-Large       & 70.7 & 69.7 & 82.0 & 93.9 & 2.7 & 0.0 & 2.9 & 0.0 & 28.6 & 18.2 & 8.8 & 3.0 & 66.2 & 57.6 & 13.2 & 6.1 & 96.8 & 100.0 & 95.3 & 93.9 & \cellcolor{blue!5}90.0 & \cellcolor{blue!5}93.9 \\
            FLUX.1-schnell       & 67.6 & 69.7 & 77.3 & 79.8 & 0.7 & 0.0 & 0.4 & 0.0 & 29.0 & 28.3 & 16.4 & 10.1 & 53.8 & 48.0 & 21.0 & 16.2 & 78.6 & 74.2 & 93.2 & 95.5 & \cellcolor{blue!5}83.7 & \cellcolor{blue!5}91.4 \\
            FLUX.1-dev       & 68.3 & 65.2 & 80.5 & 81.9 & 1.5 & 0.7 & 0.4 & 0.7 & 35.2 & 36.2 & 23.0 & 21.7 & 51.1 & 50.7 & 21.2 & 21.0 & 54.3 & 39.1 & 87.2 & 89.1 & \cellcolor{blue!5}86.5 & \cellcolor{blue!5}86.9 \\
            FLUX.1-Krea-dev       & 65.3 & 63.3 & 86.5 & 89.3 & 0.8 & 0.4 & 0.6 & 0.4 & 42.2 & 47.4 & 8.4 & 3.7 & 20.3 & 23.0 & 12.8 & 9.6 & 24.1 & 18.9 & 23.1 & 18.5 & \cellcolor{blue!5}64.2 & \cellcolor{blue!5}62.5 \\
            PixArt-$\alpha$       & 66.1 & 57.0 & 81.2 & 83.0 & 5.9 & 5.0 & 5.8 & 8.0 & 58.8 & 59.0 & 52.2 & 49.0 & 93.5 & 95.0 & 12.9 & 10.0 & 99.8 & 99.0 & 99.0 & 99.0 & \cellcolor{blue!5}99.5 & \cellcolor{blue!5}100.0 \\
            PixArt-$\Sigma$       & 64.7 & 61.9 & 89.9 & 94.3 & 3.0 & 1.0 & 2.0 & 0.0 & 44.3 & 43.8 & 30.5 & 28.6 & 88.7 & 87.6 & 14.4 & 14.3 & 99.8 & 99.1 & 97.4 & 97.1 & \cellcolor{blue!5}99.7 & \cellcolor{blue!5}100.0 \\
            HiDream-I1       & 74.0 & 72.8 & 76.2 & 85.5 & 0.7 & 0.9 & 0.4 & 0.0 & 25.8 & 31.1 & 14.8 & 15.3 & 77.6 & 82.1 & 11.2 & 11.9 & 81.1 & 79.6 & 81.6 & 82.1 & \cellcolor{blue!5}88.0 & \cellcolor{blue!5}84.6 \\
            Qwen-Image       & 53.1 & 42.3 & 87.5 & 94.2 & 0.2 & 0.0 & 0.5 & 0.0 & 37.5 & 48.1 & 1.0 & 1.9 & 48.3 & 38.5 & 19.3 & 17.3 & 57.9 & 40.4 & 90.8 & 94.2 & \cellcolor{blue!5}86.2 & \cellcolor{blue!5}86.5 \\            
            \midrule
            \multicolumn{23}{c}{\textit{Autoregressive Models}} \\
            \midrule
            Infinity-8B      & 67.3 & 67.0 & 66.7 & 71.0 & 34.3 & 20.0 & 33.0 & 20.0 & 39.7 & 43.0 & 6.2 & 1.0 & 57.0 & 66.0 & 21.8 & 16.0 & 99.3 & 96.0 & 98.8 & 98.0 & \cellcolor{blue!5}99.8 & \cellcolor{blue!5}100.0 \\
            GoT-R1-7B        & 57.6 & 75.5 & 84.5 & 91.8 & 84.1 & 67.4 & 49.1 & 28.6 & 59.1 & 61.2 & 29.2 & 16.3 & 15.5 & 6.1 & 46.0 & 34.7 & 96.3 & 95.9 & 90.8 & 85.7 & \cellcolor{blue!5}99.9 & \cellcolor{blue!5}100.0 \\
            
            \midrule
            \multicolumn{23}{c}{\textit{Unified Models}} \\
            \midrule
            BAGEL       & 71.7 & 69.0 & 80.1 & 83.0 & 2.7 & 2.0 & 1.5 & 1.0 & 39.7 & 36.0 & 16.1 & 18.0 & 57.5 & 57.0 & 19.9 & 24.0 & 88.1 & 89.0 & 93.6 & 97.0 & \cellcolor{blue!5}90.3 & \cellcolor{blue!5}92.0 \\
            BAGEL w/Think       & 67.9 & 67.8 & 84.1 & 90.1 & 3.0 & 0.8 & 0.9 & 0.8 & 37.0 & 30.6 & 21.5 & 22.3 & 61.1 & 59.5 & 21.8 & 28.9 & 90.8 & 87.6 & 96.3 & 98.4 & \cellcolor{blue!5}96.7 & \cellcolor{blue!5}98.3 \\
            show-o2-1.5B       & 56.4 & 71.4 & 85.0 & 100.0 & 22.4 & 0.0 & 15.4 & 14.3 & 20.6 & 42.9 & 14.7 & 42.9 & 88.8 & 85.7 & 35.0 & 14.3 & 100.0 & 100.0 & 97.4 & 100.0 & \cellcolor{blue!5}99.9 & \cellcolor{blue!5}100.0 \\
            show-o2-7B       & 39.2 & 53.3 & 87.9 & 86.7 & 86.2 & 76.7 & 53.6 & 46.7 & 38.4 & 43.3 & 24.5 & 13.3 & 92.6 & 96.7 & 56.8 & 46.7 & 100.0 & 100.0 & 96.8 & 96.7 & \cellcolor{blue!5}99.9 & \cellcolor{blue!5}100.0 \\
            Janus-Pro-1B       & 67.9 & 76.2 & 86.0 & 85.7 & 72.0 & 66.7 & 44.9 & 47.6 & 47.3 & 76.2 & 38.3 & 19.1 & 14.9 & 9.5 & 55.6 & 9.5 & 91.0 & 100.0 & 82.1 & 85.7 & \cellcolor{blue!5}89.3 & \cellcolor{blue!5}95.2 \\
            Janus-Pro-7B       & 69.2 & 63.3 & 84.4 & 93.3 & 71.3 & 46.7 & 43.4 & 33.3 & 49.5 & 73.3 & 27.4 & 20.0 & 15.9 & 13.3 & 52.6 & 43.3 & 93.4 & 90.0 & 88.4 & 93.3 & \cellcolor{blue!5}97.5 & \cellcolor{blue!5}93.3 \\
            BLIP3o-4B       & 60.5 & 54.5 & 73.8 & 72.3 & 31.3 & 12.9 & 46.8 & 42.6 & 64.1 & 62.4 & 16.9 & 16.8 & 83.5 & 80.2 & 16.6 & 16.8 & 99.3 & 100.0 & 99.6 & 100.0 & \cellcolor{blue!5}97.8 & \cellcolor{blue!5}98.0 \\
            BLIP3o-8B       & 65.3 & 65.1 & 73.5 & 78.6 & 39.4 & 21.4 & 51.4 & 49.5 & 70.0 & 78.6 & 19.8 & 20.4 & 76.8 & 70.9 & 15.3 & 11.7 & 99.4 & 100.0 & 99.1 & 99.0 & \cellcolor{blue!5}97.4 & \cellcolor{blue!5}96.1 \\
            OmniGen2-7B       & 66.8 & 55.2 & 76.4 & 82.8 & 7.0 & 6.9 & 5.8 & 5.2 & 33.4 & 37.9 & 34.8 & 27.6 & 65.7 & 77.6 & 35.4 & 32.8 & 97.4 & 94.8 & 93.1 & 96.6 & \cellcolor{blue!5}84.1 & \cellcolor{blue!5}98.2 \\

            \midrule
            \multicolumn{23}{c}{\textit{Closed-Source Models}} \\
            \midrule
            Seedream 3.0       & 63.0 & 50.0 & 76.0 & 77.3 & 2.9 & 0.0 & 7.1 & 4.6 & 35.5 & 47.7 & 40.9 & 64.8 & 92.4 & 92.1 & 19.6 & 10.2 & 81.9 & 75.0 & 94.3 & 92.1 & \cellcolor{blue!5}84.1 & \cellcolor{blue!5}78.4 \\
            Gemini 2.0 Flash       & 45.6 & 50.4 & 85.5 & 86.9 & 11.0 & 7.3 & 7.7 & 6.9 & 39.4 & 36.5 & 25.8 & 15.0 & 78.5 & 77.4 & 24.3 & 22.6 & 90.1 & 83.9 & 86.6 & 83.2 & \cellcolor{blue!5}79.2 & \cellcolor{blue!5}79.5 \\
            Nano Banana       & 73.9 & 80.9 & 72.3 & 78.3 & 0.5 & 0.7 & 0.4 & 1.3 & 25.5 & 20.4 & 2.3 & 1.3 & 83.3 & 81.6 & 14.4 & 18.4 & 50.8 & 41.5 & 61.9 & 57.9 & \cellcolor{blue!5}70.7 & \cellcolor{blue!5}64.4 \\
            Imagen 4       & 73.2 & 78.3 & 70.8 & 74.6 & 0.9 & 0.0 & 0.2 & 0.0 & 33.6 & 33.3 & 11.4 & 6.7 & 75.7 & 70.4 & 13.3 & 22.1 & 60.1 & 64.2 & 80.8 & 62.9 & \cellcolor{blue!5}63.5 & \cellcolor{blue!5}52.5 \\
            HunyuanImage-3.0        & 82.8 & 82.7 & 70.8 & 70.2 & 6.0 & 1.4 & 10.2 & 5.3 & 15.1 & 22.1 & 0.1 & 0.0 & 93.3 & 82.7 & 3.8 & 5.8 & 99.2 & 97.1 & 98.9 & 95.7 & \cellcolor{blue!5}97.0 & \cellcolor{blue!5}89.4 \\            
            GPT-Image       & 75.3 & 81.7 & 74.2 & 71.7 & 0.3 & 0.4 & 0.1 & 0.0 & 45.3 & 55.8 & 0.1 & 0.0 & 2.5 & 1.3 & 3.6 & 7.9 & 18.1 & 22.1 & 63.1 & 45.8 & \cellcolor{blue!5}92.1 & \cellcolor{blue!5}81.5 \\            
            \midrule
            \textbf{Average} & 
            65.3 & 66.2 & 80.0 & 83.6 & 18.3 & 12.6 & 14.4 & 11.7 & 39.1 & 43.6 & 19.7 & 17.5 & 62.0 & 60.5 & 22.6 & 17.8 & 82.9 & 80.8 & \underline{88.1} & \underline{86.9} & \cellcolor{blue!5}\textbf{89.6} & \cellcolor{blue!5}\textbf{89.5} \\
            \bottomrule
        \end{tabular}
        }
        \vspace{-3mm}
\end{table*}

\section{Experiments}
\label{sec:experiments}


\subsection{Experimental Setup}
\textbf{Datasets.} We evaluate the out-of-distribution (OOD) detection performance of MIRROR on a comprehensive set of benchmarks. To avoid shortcut learning caused by formatting discrepancies, all datasets are verified following the DDA protocol~\cite{chen2025dual}, ensuring no systematic differences in image format between real and fake samples.
\textbf{(1) Standard Benchmarks:} We evaluate on \textit{AIGCDetect}~\cite{zhong2023patchcraft}, \textit{GenImage}~\cite{zhu2023genimage}, \textit{DRCT-2M}~\cite{chen2024drct}, \textit{UnivFakeDetect}~\cite{ojha2023towards}, \textit{Synthbuster}~\cite{bammey2023synthbuster}, and \textit{EvalGEN}~\cite{chen2025dual}. These benchmarks cover dozens of generative models spanning GANs, diffusion models, and autoregressive architectures.
\textbf{(2) In-the-wild Benchmarks:} To assess real-world generalization, we use \textit{Chameleon}~\cite{yan2024sanity}, \textit{SynthWildx}~\cite{cozzolino2024raising}, \textit{WildRF}~\cite{cavia2024real}, \textit{AIGIBench}~\cite{li2025artificial}, \textit{CO-SPY-Bench}~\cite{cheng2025co}, \textit{BFree-Online}~\cite{guillaro2025bias}, and \textit{RRDataset}~\cite{li2025bridging}. These datasets are collected from complex network environments and include realistic degradations introduced during online propagation.
\textbf{(3) Human-AIGI Benchmark:} Crucially, we evaluate MIRROR on $\mathcal{D}_{orig.}$ and $\mathcal{D}_{hard}$, defined in Sec.~\ref{sec:human_benchmark}, to measure performance in high-risk scenarios where human visual inspection fails.

\textbf{Implementation Details.} In Phase~1, we adopt DINOv3-large~\cite{simeoni2025dinov3} as the feature extractor and collect 200k real images from MSCOCO~\cite{lin2014microsoft} for self-supervised training. The memory bank size is set to $K=4096$.
In Phase~2, we use a LoRA-finetuned DINOv3 as the feature extractor and finetune the model on SDv1.4~\cite{zhu2023genimage} \footnote{The SD v1.4 images from GenImage used in the training set are compressed using JPEG with a quality factor of 96 to align their format with that of real images.}. Training is performed with the AdamW optimizer~\cite{kingma2014adam}, an initial learning rate of $1\times10^{-4}$, for 5 epochs, using a cosine annealing learning rate schedule. All images are cropped to a resolution of $224 \times 224$ during both training and inference.

\begin{figure}[t!]
    \centering
    \includegraphics[scale=0.285]{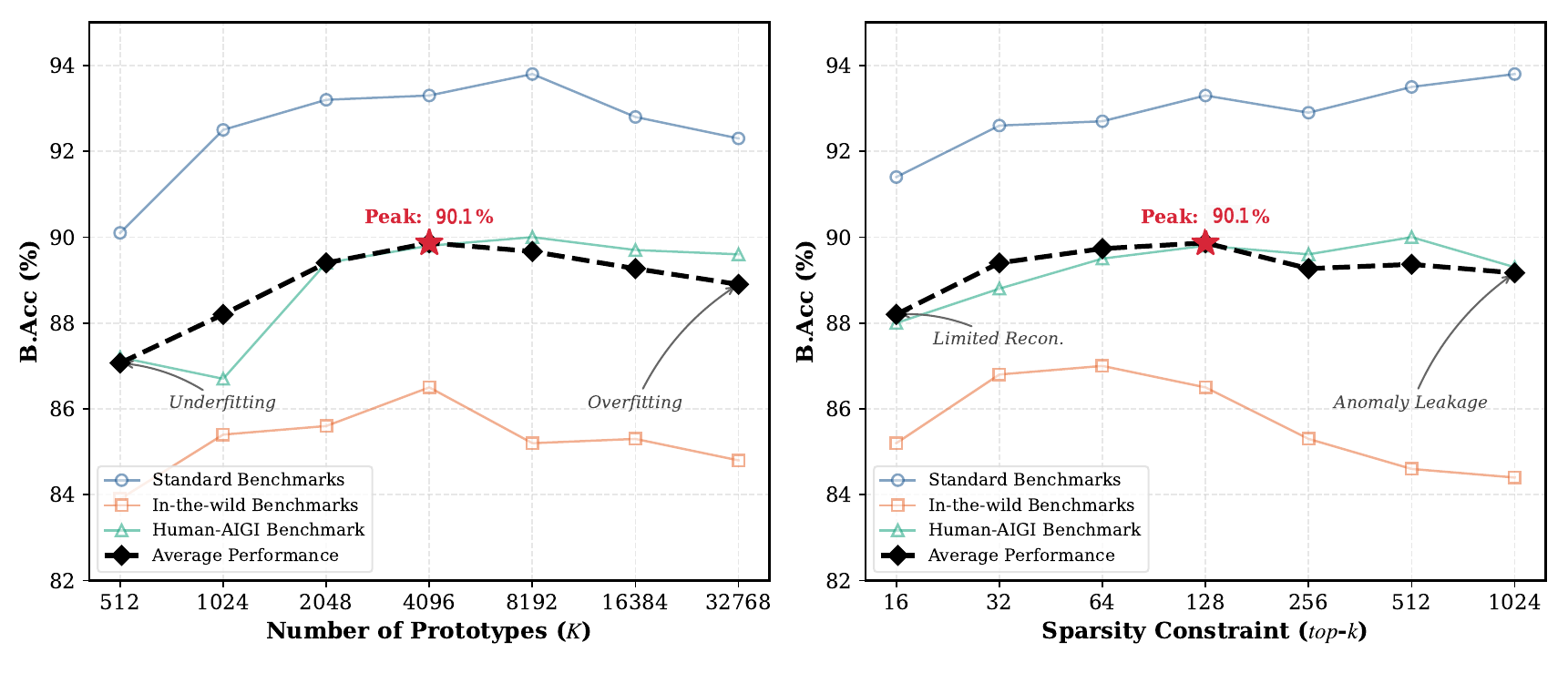}
\caption{\textbf{Sensitivity analysis of $K$ and $top$-$k$.} Insufficient $K$ limits manifold coverage, while excessive $K$ causes overfitting via prototype redundancy. Sparsity $top$-$k$ exhibits a similar trade-off: strict $k$ hinders ideal reference reconstruction , while excessive $k$ allows generative anomalies to leak through prototype combinations, narrowing the residual gap.}
    \label{fig:mmbank}
    \vspace{-5mm}
\end{figure}

\begin{table}[t!]
    \centering
    \caption{\textbf{Ablation Studies on MIRROR Architecture.} We analyze the impact of different configurations on detection performance. \textbf{(a)} Compares different data sources for constructing the Memory Bank. \textbf{(b)} Validates the necessity of the Reference-Comparison paradigm compared to a discriminative baseline.}
    \label{tab:ablation_combined}
    
    \resizebox{0.98\linewidth}{!}{
        \begin{tabular}{l | c c c | c}
            \toprule
            \textbf{Configuration} & \textbf{Standard} & \textbf{In-the-wild} & \textbf{Human-AIGI} & \textbf{Average} \\
            \midrule
            
            \multicolumn{3}{l}{\textit{\textbf{(a) Memory Prior Source}}} \\
            \midrule
            Generated-only & 90.5 & 84.4 & 84.9 & 86.6\\
            Mixed (Real + Generated) & 87.1 & 82.3  & 84.3 & 84.5\\
            \rowcolor{blue!5}
            \textbf{Real-only (Ours)} & \textbf{94.0} & \textbf{86.5}  & \textbf{89.8} & \textbf{90.1}\\
            \midrule
            
            \multicolumn{3}{l}{\textit{\textbf{(b) Reference-Comparison}}} \\
            \midrule
            Baseline (Only Classify) & 88.4 & 82.2 & 83.7 & 84.7\\
            + Perplexity ($V_{per}$) & 90.1 & 84.3 & 85.0 & 86.4\\
            + Residual ($V_{res}$) & 92.5 & 85.1 & 88.5 & 88.7\\
            \rowcolor{blue!5}
            \textbf{MIRROR (Ours)} & \textbf{94.0} & \textbf{86.5} & \textbf{89.8} & \textbf{90.1}\\
            \bottomrule
        \end{tabular}
    }
    \vspace{-3mm}
\end{table}


\textbf{Evaluation Metrics and Comparative Methods.}
We report three evaluation metrics: Balanced Accuracy (B.Acc), JPEG Robustness (J.Rob) under QF=90 compression, and Resize Robustness (R.Rob) under resize=0.9. We compare MIRROR against a broad set of state-of-the-art detectors.
Artifact- and pretraining-based methods include NPR~\cite{tan2024rethinking}, UnivFD~\cite{ojha2023towards}, FatFormer~\cite{liu2024forgery}, C2P-CLIP~\cite{tan2025c2p}, SAFE~\cite{li2024improving}, and AIDE~\cite{yan2024sanity}. These methods capture generic forgery traces through architectural design or specialized training. For fair comparison, all methods are retrained using the same data configuration as MIRROR, including the real samples used in Phase~1.
We also evaluate dataset-alignment-based detectors, including DRCT~\cite{chen2024drct}, Aligned~\cite{rajan2024aligned}, B-Free~\cite{guillaro2025bias}, and DDA~\cite{chen2025dual}. These approaches mitigate semantic or formatting biases to isolate forgery-related cues. For methods that rely on alignment datasets, we evaluate them using their official pretrained weights.

\subsection{Comparison with State-of-the-Art}

\textbf{Generalization on Standard Benchmarks.}
As shown in Table~\ref{tab:standard_benchmarks}, we evaluate MIRROR on five standard datasets spanning diverse generative architectures. MIRROR consistently outperforms state-of-the-art methods, especially in cross-generator settings, achieving an average Balanced Accuracy (B.Acc) of \textbf{94.0\%}, substantially surpassing AIDE~\cite{yan2024sanity}. This indicates that leveraging robust priors learned from real-world data effectively mitigates overfitting to the training distribution. While alignment-based methods such as B-Free~\cite{guillaro2025bias} attain competitive performance by reducing semantic bias, they exhibit sensitivity to mild compression. In contrast, MIRROR’s expectation-deviation paradigm yields stronger robustness, outperforming B-Free by \textbf{2.0\%} / \textbf{3.5\%} in JPEG Robustness (J.Rob) / Resize Robustness (R.Rob) .

\textbf{Generalization on In-the-Wild Benchmarks.} 
Real-world deployment is challenged by image degradations introduced during internet propagation. As reported in Table~\ref{tab:wild}, we evaluate MIRROR on seven in-the-wild benchmarks. Detectors without pretrained backbones (e.g., NPR~\cite{tan2024rethinking}, SAFE~\cite{li2024improving}) fail to generalize, while pretraining-based and alignment-based methods provide only limited gains. In contrast, MIRROR achieves the strongest overall robustness, maintaining a B.Acc of \textbf{91.2\%} across diverse post-processing conditions and outperforming DDA~\cite{chen2025dual} by 8.1\%. Notably, unlike DDA, which relies on large-scale real and VAE-reconstructed datasets, MIRROR requires only readily available real images for prior learning and a small amount of generated data for fine-tuning.

\textbf{Detection of Human-Imperceptible Threats.}
A critical evaluation of our framework lies in comparing its performance with that of humans, particularly on subsets that are imperceptible to human observers (see Sec.~\ref{sec:human_benchmark}).As reported in Table~\ref{tab:wild}, We observe that detectors relying on low-level cues perform poorly across all generators, while detectors based on pretrained priors achieve only limited improvements and still struggle on the human-imperceptible subset. In contrast, \textbf{MIRROR} achieves a balanced accuracy of up to \textbf{89.6\%} on the original dataset, surpassing both lay users (76.9\%) and visual experts (88.3\%). Moreover, MIRROR maintains a high balanced accuracy of \textbf{89.5\%} on the human-imperceptible subset. These results highlight the core value of our approach, demonstrating its effectiveness as an assistive tool for detecting generative threats that evade human perceptual scrutiny.

\subsection{Ablation Study and Analysis}
\label{sec:ablation}

\textbf{Impact of Capacity $K$ and Sparsity $top$-$k$.}
As shown in Fig.~\ref{fig:mmbank}, performance follows an inverted U-shaped trend with respect to $K$, peaking at $4096$. Insufficient capacity fails to cover the real-image manifold, while excessive capacity introduces redundancy and overfitting. A similar trade-off appears for projection sparsity $top$-$k$: small values underfit complex textures, whereas large values dilute residual separability by absorbing generative artifacts. We therefore set $K=4096$ and $top$-$k=128$ as a balanced configuration.

\textbf{Impact of Memory Prior.} We compare Memory Bank prototypes sourced from real, generated, or mixed samples. As shown in Table~\ref{tab:ablation_combined} (a), the real-only configuration performs best, while generated-only slightly degrades due to the lower distributional consistency of forgeries. In contrast, the mixed setting leads to a severe performance drop, as combining authentic and forged prototypes breaks the Reference–Comparison paradigm, causing residuals to lose discriminative power.

\begin{figure}[t]
    \centering
    \includegraphics[scale=0.27]{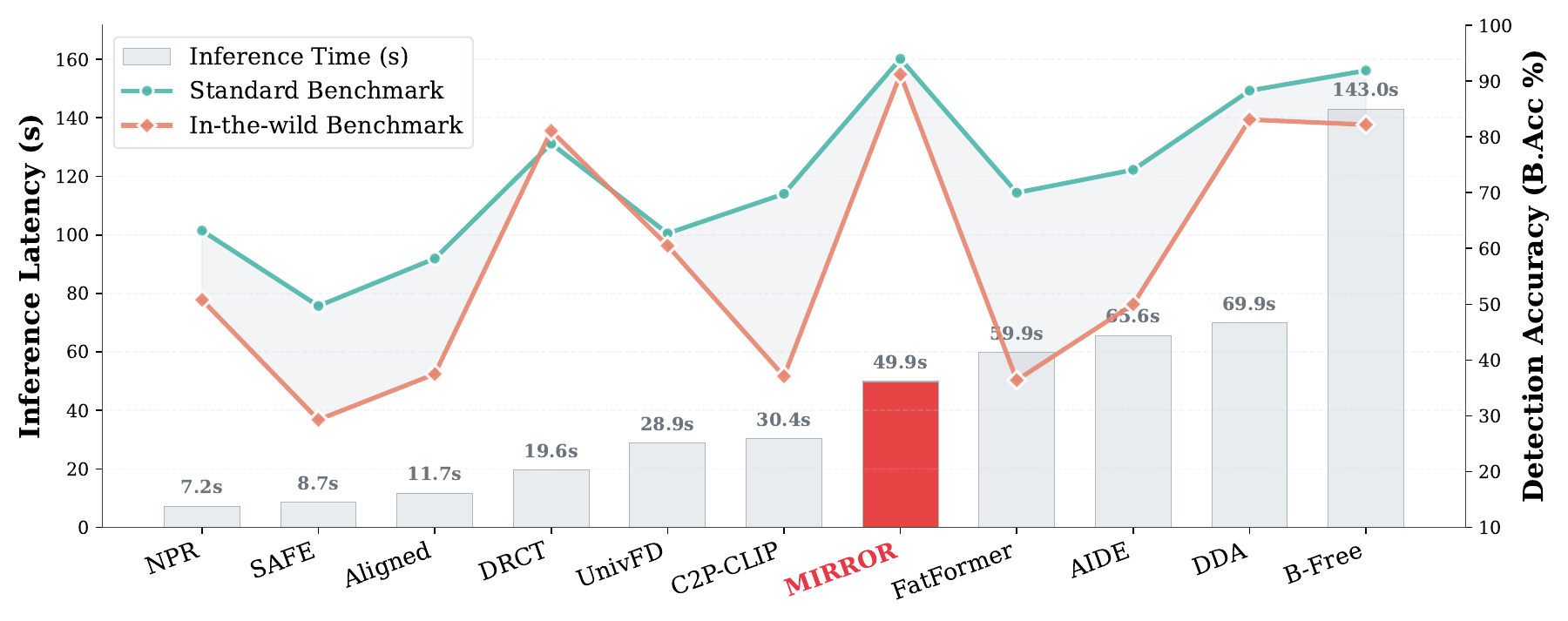}
\caption{\textbf{Inference efficiency comparison.} Time per 1,000 samples on an Nvidia Tesla V100 GPU. MIRROR's inference speed is highly competitive among existing detectors.}
    \label{fig:time}
    \vspace{-3mm}
\end{figure}

\begin{figure}[t]
    \centering
    \includegraphics[scale=0.072]{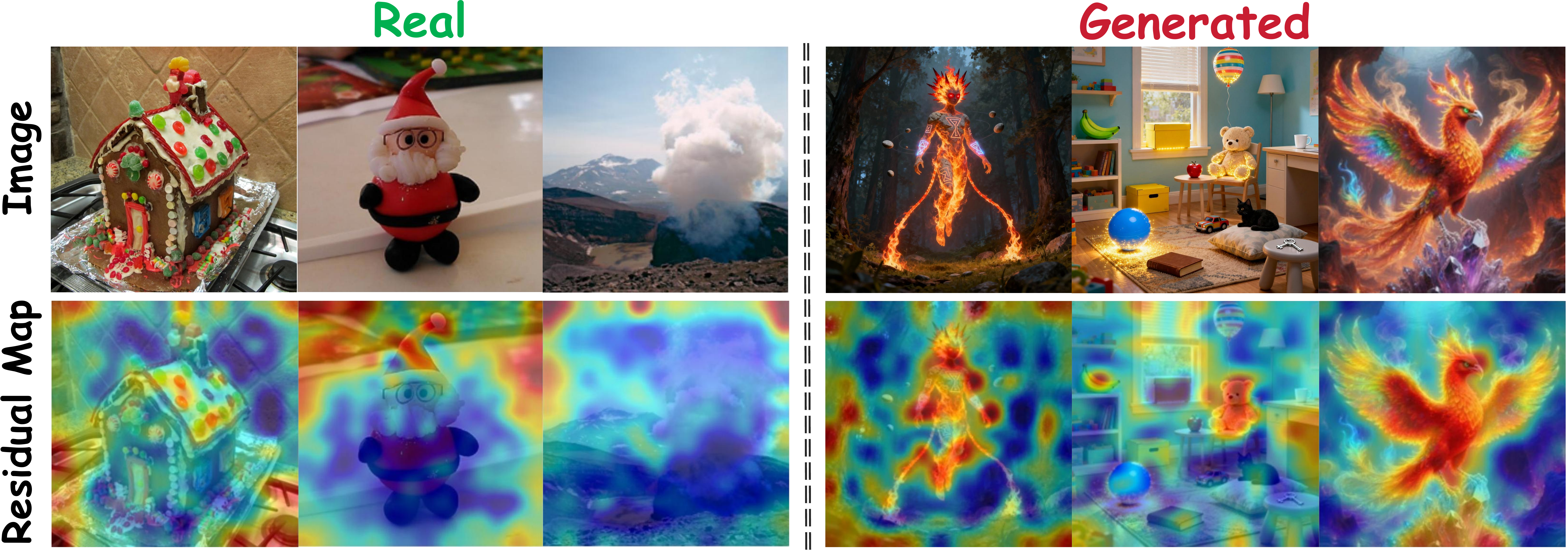}
\caption{\textbf{Feature reconstruction residuals.} Real images produce sparse, low-magnitude residuals due to effective manifold projection, while generated images show pronounced residual activations in regions that violate internal priors.}
    \label{fig:vis}
    \vspace{-4mm}
\end{figure}

\textbf{Impact of the Reference-Comparison Paradigm.}
To validate the advantage of Reference–Comparison over black-box classification, we compare MIRROR with a baseline that directly classifies encoded features. As shown in Table~\ref{tab:ablation_combined}(b), the baseline suffers a \textbf{5.4\%} drop relative to MIRROR. Moreover, progressively adding Reconstruct Perplexity and Comparison Residual consistently improves performance, confirming the effectiveness of the human-inspired Reference–Comparison paradigm for robust generalization.

\textbf{Visualization of Reconstruction Residuals.}
Fig.~\ref{fig:vis} shows feature reconstructions for real and generated images. For real samples, reconstructed features $\hat{F}$ closely match input features $F$, yielding sparse, low-magnitude residuals overall. In contrast, generated images produce pronounced residual activations, typically concentrated in semantically complex regions. This pattern reflects human-like reference-based verification and provides interpretable evidence for the model’s decisions.

\textbf{Computational Efficiency.}
We evaluate MIRROR’s inference efficiency relative to standard detectors. Despite the discrete memory bank and reference  reconstructor, the overhead is minimal since inference is dominated by the DINOv3-large backbone. On a single Nvidia Tesla V100 GPU, MIRROR achieves 20.03 FPS, processing 1,000 samples in 49.91 seconds. As shown in Fig.~\ref{fig:time}, this performance is competitive with black-box detectors, demonstrating that MIRROR framework maintains deployment scalability while improving generalization.

%% file: sections/05_conclusion.tex
\section{Conclusion}
We identify a fundamental limitation of artifact-driven AIGI detectors, namely their poor scalability and generalization. Inspired by human visual verification, we propose MIRROR, which formulates detection as a Reference–Comparison process by modeling the real-image manifold with a learnable memory bank. Extensive experiments, including the proposed Human-AIGI benchmark, show that MIRROR consistently outperforms existing methods and surpasses human performance on human-imperceptible samples.